\documentclass[letterpaper, 10 pt, conference]{ieeeconf}

\IEEEoverridecommandlockouts
\usepackage{graphicx}
\usepackage{color}
\usepackage[cmex10]{amsmath}
\usepackage{amssymb}
\usepackage[tight,footnotesize]{subfigure}
\usepackage{float}
\usepackage{multirow}
\usepackage{booktabs}
\usepackage{subeqnarray}
\usepackage{enumerate}
\usepackage{epstopdf}
\usepackage{wrapfig}
\usepackage{soul}
\usepackage[table]{xcolor}
\usepackage{array}
\usepackage{algorithm}
\usepackage[noend]{algorithmic}
\usepackage{comment}

\usepackage{bbm}

\usepackage{cite}
\usepackage[bookmarks=true,colorlinks=true,linkcolor=black,backref=true,
pdfborder={0 0 0},citecolor=blue,urlcolor=blue]{hyperref}

\makeatletter
\let\proof\@undefined			
\let\endproof\@undefined		
\makeatother
\usepackage{amsthm}

\newtheorem{prob}{Problem}

\theoremstyle{definition}

\newcommand{\real}[1][]{\mathbb{R}^{#1}}                                
\newcommand{\nat}[1][]{\mathbb{N}^{#1}}                                 

\newcommand{\defeq}{:=}                                                 
\newcommand{\msub}[1]{_\mathrm{#1}}                                     

\newcommand{\clint}[2]{\left[#1, #2\right]}                             

\renewcommand{\leq}{\leqslant}                                          

\newcommand{\intset}[1]{0:#1}

\newcommand{\E}[2]{\mathbb{E}_{#2}\left[ #1 \right]}

\newcommand{\mydef}[1]{{\textit{#1}}}

\newcommand{\ppl}{ath-planning}

\newcommand{\matlab}{MATLAB\textsuperscript{\textregistered}}

\newcommand{\eqnnt}[1]{\hyperref[#1]{(\ref*{#1})}}
\newcommand{\eqnsnt}[2]{\hyperref[#1]{(\ref*{#1})}
	and~\hyperref[#2]{(\ref*{#2})}}
\newcommand{\eqnsernt}[2]{\hyperref[#1]{(\ref*{#1})}--\hyperref[#2]{(\ref*{#2})}}

\newcommand{\eqn}[1]{\hyperref[#1]{Eqn.~(\ref*{#1})}}
\newcommand{\eqns}[2]{\hyperref[#1]{Eqns.~(\ref*{#1})} and~\hyperref[#2]{(\ref*{#2})}}
\newcommand{\eqnser}[2]{\hyperref[#1]{Eqns.~(\ref*{#1})}--\hyperref[#2]{(\ref*{#2})}}
\newcommand{\eqnf}[1]{\hyperref[#1]{Equation~(\ref*{#1})}}
\newcommand{\eqnfs}[2]{\hyperref[#1]{Equations~(\ref*{#1})} and~\hyperref[#2]{(\ref*{#2})}}

\newcommand{\scn}[1]{\hyperref[#1]{Sec.~\ref*{#1}}}
\newcommand{\scns}[2]{\hyperref[#1]{Secs.~\ref*{#1}} and~\hyperref[#2]{\ref*{#2}}}
\newcommand{\scnser}[2]{\hyperref[#1]{Secs~.\ref*{#1}}--\hyperref[#2]{\ref*{#2}}}

\newcommand{\fig}[1]{\hyperref[#1]{Fig.~\ref*{#1}}}
\newcommand{\figs}[2]{\hyperref[#1]{Figs.~\ref*{#1}} and~\hyperref[#2]{\ref*{#2}}}
\newcommand{\figser}[2]{\hyperref[#1]{Figs.~\ref*{#1}}--\hyperref[#2]{\ref*{#2}}}
\newcommand{\figf}[1]{\hyperref[#1]{Figure~\ref*{#1}}}
\newcommand{\figfs}[2]{\hyperref[#1]{Figures~\ref*{#1}} and~\hyperref[#2]{\ref*{#2}}}
\newcommand{\figfser}[2]{\hyperref[#1]{Figures~\ref*{#1}}--\hyperref[#2]{\ref*{#2}}}

\newcommand{\tbl}[1]{\hyperref[#1]{Table~\ref*{#1}}}
\newcommand{\tbls}[2]{\hyperref[#1]{Tables~\ref*{#1}} and~\hyperref[#2]{\ref*{#2}}}
\newcommand{\tblser}[2]{\hyperref[#1]{Tables~\ref*{#1}}--\hyperref[#2]{\ref*{#2}}}

\newcommand{\apx}[1]{\hyperref[#1]{Appendix~\ref*{#1}}}

\newcommand{\prb}[1]{\hyperref[#1]{Problem~\ref*{#1}}}
\newcommand{\prp}[1]{\hyperref[#1]{Prop.~\ref*{#1}}}
\newcommand{\prpf}[1]{\hyperref[#1]{Proposition~\ref*{#1}}}

\newcommand{\algoref}[1]{\hyperref[#1]{Algorithm~\ref*{#1}}}

\newcommand{\thmref}[1]{\hyperref[#1]{Theorem~\ref*{#1}}}
\newcommand{\thmsref}[2]{\hyperref[#1]{Theorems~\ref*{#1}} and~\hyperref[#2]{\ref*{#2}}}
\newcommand{\thmserref}[2]{\hyperref[#1]{Theorems~\ref*{#1}}--\hyperref[#2]{\ref*{#2}}}

\newcommand{\algline}[1]{\hyperref[#1]{Line~\ref*{#1}}}
\newcommand{\alglines}[2]{\hyperref[#1]{Lines~\ref*{#1}} and~\hyperref[#2]{\ref*{#2}}}
\newcommand{\alglineser}[2]{\hyperref[#1]{Lines~\ref*{#1}}--\hyperref[#2]{\ref*{#2}}}

\newcommand{\tento}[1]{\times 10^{#1}}

\renewcommand{\vec}[1]{\boldsymbol{#1}}

\newcommand{\wsp}{\mathcal{W}}
\newcommand{\tf}{T}
\newcommand{\threat}{c}
\newcommand{\threatState}{\vec{\Theta}}
\newcommand{\nParam}{N\msub{P}}
\newcommand{\nGridPts}{N\msub{G}}

\newcommand{\nnParam}{\theta}
\newcommand{\nnParamWt}{w}
\newcommand{\nnParamBs}{b}
\newcommand{\nnParamEnc}{\phi}
\newcommand{\nnParamDec}{\theta}

\def\encoder{e}
\def\decoder{g}

\newcommand{\rnnState}{h}

\newcommand{\nData}{N\msub{D}}
\newcommand{\nGen}{N\msub{S}}
\newcommand{\nTime}{N\msub{T}}

\newcommand{\datum}{x}

\newcommand{\dataset}{\mathcal{X}}
\newcommand{\datasetGen}{\tilde{\mathcal{X}}}
\newcommand{\datasetSup}{\mathcal{X}\msub{s}}

\newcommand{\pos}{\vec{r}}

\newcommand{\yOutput}{y}

\newcommand{\latentSpace}{\mathcal{Z}}
\newcommand{\latentVector}{z}

\newcommand{\loss}{L}

\newcommand{\klDiv}[1]{D\msub{KL}\left(#1\right)}

\def\indicator{\vec{1}\msub{s}}

\title{\LARGE \bf
Synthetic Data Generation for Minimum-Exposure Navigation 
in a Time-Varying Environment using Generative AI Models
}

\author{Nachiket U. Bapat$^{1}$, Randy C. Paffenroth$^{2}$ and Raghvendra V. Cowlagi$^{1\dagger}$
\thanks{$^{1}$Aerospace Engineering Dept., $^{2}$Mathematical Sciences Dept.}%
\thanks{\textit{Worcester Polytechnic Institute,} 100 Institute Rd, Worcester MA USA.}
\thanks{$^{\dagger}$Corresponding author. Email: \texttt{\footnotesize rvcowlagi@wpi.edu}}
}

\begin{document}

\maketitle
\thispagestyle{empty}
\pagestyle{empty}

\begin{abstract}
	
	We study the problem of synthetic generation of samples of
	environmental features for autonomous vehicle navigation. These features
	are described by a spatiotemporally varying scalar field that we refer to 
	as a threat field. The threat field is known to have some underlying
	dynamics subject to process noise. Some ``real-world'' data of observations
	of various threat fields are also available. The assumption is that the volume
	of ``real-world'' data is relatively small. The objective is to synthesize
	samples that are statistically similar to the data. The proposed solution
	is a generative artificial intelligence model that we refer to as a 
	split variational recurrent neural network (S-VRNN). The S-VRNN merges
	the capabilities of a variational autoencoder, which is a widely used
	generative model, and a recurrent neural network, which is used to learn
	temporal dependencies in data. The main innovation in this work is that 
	we split the latent space of the S-VRNN into two subspaces.
	The latent variables in one subspace are learned using the ``real-world'' data, 
	whereas those in the other subspace are learned using the data as well as the
	known underlying system dynamics. 
	Through numerical experiments we demonstrate that the proposed S-VRNN
	can synthesize data that are statistically similar to the training data 
	even in the case of very small volume of ``real-world'' training data.
	
\end{abstract}

\section{Introduction}

Systems like autonomous mobile vehicles -- whether aerial or
terrestrial -- are expensive to operate in the 
real world. The design and validation of controllers for such systems
therefore relies on a combination of mathematical modeling, 
abundant numerical simulations, and a relatively small set of
real-world experiments.
Simulations are developed by executing mathematical models, e.g., 
integration of state-space differential equations of the system, 
to computationally synthesize data of the system's operation, 
e.g.,~\cite{Dosovitskiy2017}.

These synthetic data are essential due to the scarcity of real-world operational data.
In typical model-based control design methods, synthetic data may be used
for preliminary validation of the controller. More recent model-free reinforcement
learning (RL) methods need large volumes of synthetic data during the training phase
\cite{kiumarsi2018optimal,kuutti2021survey}.
Other deep learning (DL) methods, such as vision-based object detectors 
and classifiers widely used in various subsystems of autonomy, 
also need large volumes of training data~\cite{Gupta2022,Sisson2023,Sprockhoff2024}.

The mathematical models used for simulations encode an understanding of 
the system's behavior, e.g., geometric constraints and the laws of physics. 
Almost without exception, these models involve some simplifications, approximations, 
and epistemic uncertainties such as inexact knowledge of the system's properties. 
Aleatoric uncertainties such as environmental disturbances may also be present, 
and are sometimes approximated within the simulation model.
Nevertheless, due to all of these discrepancies, the data synthesized by 
simulation models does not match data from the system's real-world 
operation~\cite{Jakobi1995}, which is the fundamental problem of 
\emph{model mismatch.} RL-based controllers, for example, are known to suffer 
from real-world performance degradation due to a 
``reality gap''~\cite{francois2018introduction},  i.e., model mismatch.

System identification (ID)
methods alleviate this problem by tuning various parameters 
in the simulation model using 
real-world data~\cite[pp. 97 -- 155]{Jategaonkar2006}. 
A caveat are that the accuracy of system ID relies on real-world data,
the scarcity of which is the root problem.
A reduction in the mismatch between synthetic data and real-world operational 
data can potentially deliver improvements not only in control design 
and validation, but also in other areas such as performance / reliability 
analyses and digital twin development.

Recent years have witnessed explosive advances in computational data synthesis
through \mydef{generative artificial intelligence models} (GAIMs). 
Loosely speaking, a GAIM is an DL model that learns to transform a set of 
uniformly distributed
latent vectors to a set of output vectors with a distribution similar -- in the sense
of small Kullback-Liebler (KL) divergence or Wasserstein distance -- 
to that of a training dataset~\cite{nikolenko2021synthetic}. 
Well-known examples of GAIMs include GPT-4 (which underlies the 
ChatGPT chatbot application), the image generator DALL-E
\cite{ramesh2022hierarchicaltextconditionalimagegeneration}, 
the software code generator GitHub Copilot~\cite{nguyen2022empirical}, 
and the human face generator StyleGAN~\cite{melnik2024face}.

Considering the success of GAIMs in image- and natural language processing (NLP),
one may consider a broad research question of whether GAIMs may be developed 
to reduce the mismatch between synthetic and real-world data. To investigate 
this question further, note that there are two main issues where GAIM development 
for engineering systems contrasts GAIMs in the image processing
and NLP domains: (1) training data is scarce for systems of our interest, e.g.,
autonomous vehicles moving in an adversarial threat field,
and (2) these systems are governed by underlying
physical and algorithmic principles, namely, natural laws and control laws.

In this paper we study the problem of synthetic generation of samples of
environmental features for autonomous vehicle navigation. These features
are described by a spatiotemporally varying scalar field that we refer to 
as a \mydef{threat field.} The threat field is known to have some underlying
dynamics subject to process noise. Some ``real-world'' data of observations
of various threat fields are also available. The assumption is that the volume
of ``real-world'' data is relatively small. The objective is to synthesize
samples that are statistically similar to the data. The proposed solution
is a GAIM that we refer to as a split variational recurrent neural network.
Whereas the eventual goal is to use these synthetic data to develop autonomous 
p\ppl\ algorithms for minimizing exposure to the threat, we defer the 
p\ppl\ problem to future work.

\textit{Related Work:} Two of the most widely used GAIM architectures are generative
adversarial networks (GANs)~\cite{goodfellow2014generative}  and 
variational autoencoders (VAEs)~\cite{kingma2019introduction-vae}. The proposed
work is related to VAEs, which we briefly describe in \scn{ssec-vae}.
Recent literature explores GAIMs with temporal dependencies in data.
For example, the TimeGAN~\cite{yoon2019time-series-gan}  
addresses the temporal correlations between data points in time series,
and reports on experiments with datasets including multivariate sinusoidal
signals, stock prices, and energy consumption patterns. The Quant GAN~\cite{wiese2020quant-gans}
utilizes convolutional neural networks (CNNs) to analyze financial data. 
Similarly, the Convolutional LSTM approach~\cite{NIPS2015_07563a3f} merges the 
capabilities of CNNs and Long Short-Term Memory (LSTM) networks designed 
to process spatiotemporal data such as video sequences and weather data.
Variational Recurrent Neural Networks (VRNNs)~\cite{fabius2015variationalrecurrentautoencoders} 
address temporal dependencies during the generation process by integrating 
VAE with a recurrent neural network (RNN) at each time step. VRNNs demonstrate
effective performance in tasks such as speech modeling and handwriting 
generation~\cite{nguye2020}. The thesis~\cite{nguye2020} presents a geospatial 
detector for anomaly detection using leverages variational deep learning.

The proposed GAIM is a novel modification of the VRNN architecture. 
Like any other DL model, the VRNN encodes its inputs in a hidden layer variables,
called latent variables.
The main innovation in this work is that we split the latent space of 
the model into two subspaces.
The latent variables in one subspace are learned using the ``real-world'' data, 
whereas those in the other subspace are learned using the data as well as the
known underlying system dynamics. This is achieved by augmenting the ``real-world'' 
dataset with data synthesized via noiseless simulation of the system. 
Through numerical experiments we demonstrate that the proposed model
is able to synthesize time-series data that are statistically similar 
to the training data even in the case of very small volume of ``real-world''
training data.

The rest of this paper is organized as follows.
In \scn{Prob}, we introduce the problem of interest. 
In \scn{NN_architecture}, we describe
the proposed generative model architecture. 
In \scn{sec-results}, we provide results of the proposed synthetic 
data generating method, and conclude the paper in \scn{sec-conclusions}.

\section{Problem Statement}
\label{Prob}

We summarize commonly used notation in \tbl{tbl-notation}.
\begin{table}
	\centering
	\caption{Commonly used notation in this paper.}%
	\label{tbl-notation}.
	\begin{tabular}{p{0.025\columnwidth} p{0.25\columnwidth} | p{0.07\columnwidth} 
	p{0.45\columnwidth}}
		\toprule
		$\nat$ & Natural numbers & $\intset{n}$ & $\{0, 1, \ldots, n\}$ for $n \in \nat$ \\
		\midrule
		$\real$ & Real numbers & $\real[n]$ & Real vector space of dim. $n \in \nat$ \\
		\midrule
		\multicolumn{4}{l}{
		$\mathcal{N}(\mu, \Sigma)$ \qquad ~Normal distribution with mean $\mu$ and covariance 
		$\Sigma$} \\
		\midrule
		\multicolumn{4}{l}{$\E{f(x)}{x}$ \qquad Expectation of $f(x)$ over $x$} \\ \midrule
		\multicolumn{4}{l}{$\klDiv{q \parallel p}$ ~~Kullback-Leibler divergence from $q$ to $p$} \\
		\bottomrule
	\end{tabular}
\end{table}

Consider a compact square 2D environment $\wsp \subset \real[2]$ and a finite
time interval $\clint{0}{\tf}.$
We are interested in the problem of generating samples of a spatiotemporally-varying
threat field $\threat: \wsp \times \clint{0}{\tf} \rightarrow \real.$
We restrict this study to threat fields that can be expressed using a separation
of spatial and temporal variables as:
\begin{align}
	\threat(\pos, t) &= 1 + \vec{\Phi}^\intercal(\pos) \threatState(t),
	\quad \mbox{for } \pos \in \wsp, ~t \in \clint{0}{\tf}.
	\label{eq-threat}
\end{align}
Here $\vec{\Phi} := \left(\varphi_1(\boldsymbol{r}), 
\varphi_2(\boldsymbol{r}), \dots \varphi_{\nParam}(\boldsymbol{r})\right)^\intercal$, 
is a spatial basis function vector, for some prespecified $\nParam \in \nat.$ 
We choose the radial basis functions
\begin{align*}
	\varphi_i(\boldsymbol{r}) := \exp\left(-\frac{(\boldsymbol{r} - 
	\vec{b}_i)^\intercal (\boldsymbol{r} - \vec{b}_i)}{2a_i}\right),
\end{align*}
for $i \in \intset{\nParam}.$ The parameters  $a_i \in \real_{>0}$ 
and $\vec{b}_i \in \wsp$ are chosen arbitrarily.
We assume that the time-varying coefficient vector $\threatState(t)$ is
governed by underlying linear dynamics of the form
\begin{align}
	\dot{\threatState}(t) = A \threatState(t) + \eta_1(t),
	\label{eq-dynamics}
\end{align}
where $\eta_1$ is process noise. $A \in \real[\nParam \times \nParam]$
is assumed to be known and Hurwitz.
Crucially, we make no specific assumptions are made regarding
the process noise characteristics. In other words, knowledge of the
process- (and measurement-) noise is available implicitly through
data, only.

The objective is to synthesize samples of such a threat field such that 
the synthesized samples are statistically similar to a dataset of threat fields.
Typical forward-integrating simulations 
cannot be used for this purpose due to the lack of any knowledge of the 
characteristics of~$\eta_1.$

\begin{figure}
	\centering
	\subfigure[$t_1 = 1.$]{\includegraphics[width=0.46\columnwidth]{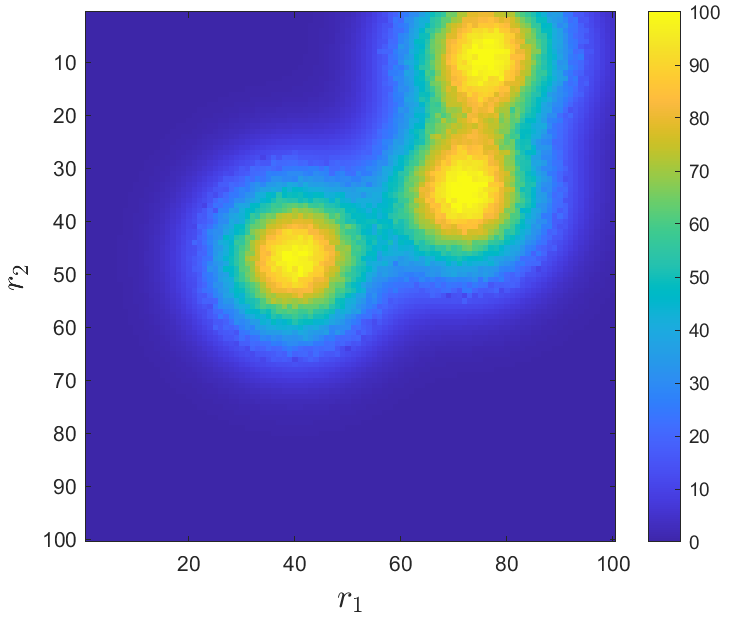}}
	\hspace{0.02\columnwidth}
	\subfigure[$t_2 = 2.$]{\includegraphics[width=0.46\columnwidth]{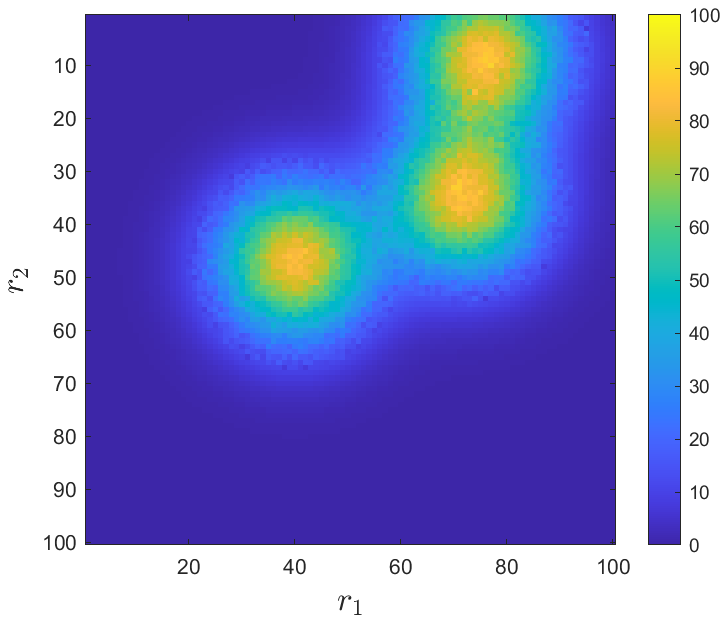}}
	\hspace{0.02\columnwidth}
	\subfigure[$t_3 = 3.$]{\includegraphics[width=0.46\columnwidth]{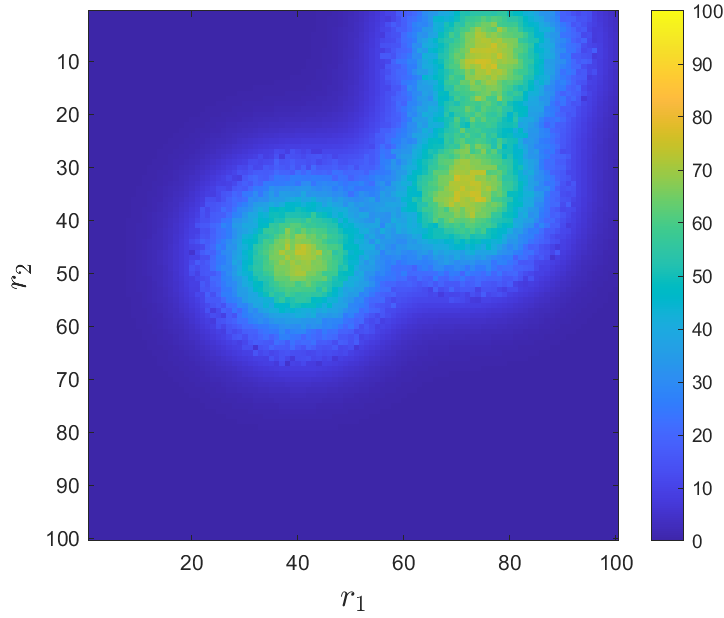}}
	\hspace{0.02\columnwidth}
	\subfigure[$t_4 = 4.$]{\includegraphics[width=0.46\columnwidth]{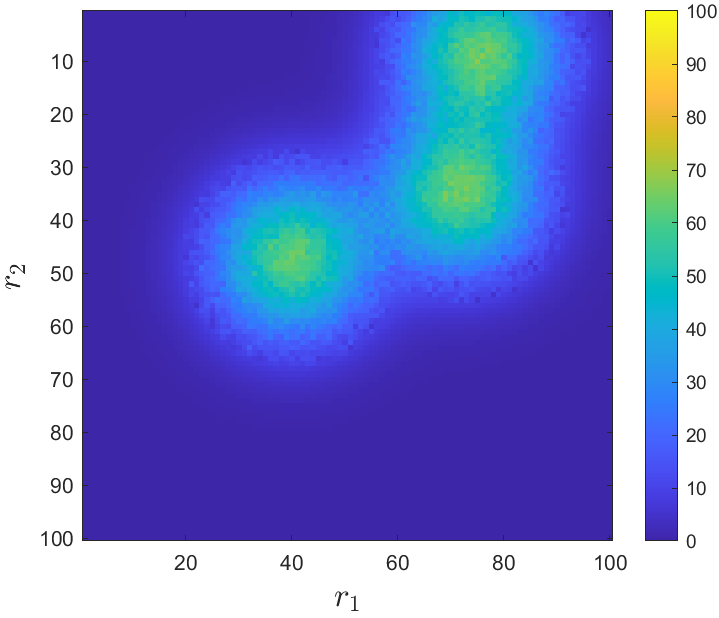}}
	\caption{Sample threat field representing a single data point.}
	
	\label{fig-real_threat}
\end{figure}

The ``real-world'' dataset of threat fields is a set of measurements
of various threat fields at various points of time. We assume that the measured
quantity is the threat intensity values over a spatial grid of points,
with additive measurement noise $\eta_2(t).$ Specifically, consider a set of
spatial grid points $\pos_{1}, \ldots, \pos_{\nGridPts} \in \wsp,$ for some
$\nGridPts \in \nat.$ An observation vector at time $t$ is
\begin{align}
	\datum_t &= \left( \threat(\pos_{1}, t), \threat(\pos_{2}, t), \ldots,
		\threat(\pos_{\nGridPts}, t) \right) + \eta_2(t).
	\label{eq-observation}
\end{align}
In a minor abuse of notation, henceforth we use $t$ to denote the \emph{index} 
of a time step rather than the absolute time.
A \mydef{datum} or \mydef{data point}) is a time-series of observations,
i.e., $\datum = \{\datum_t\}_{t \in \intset{T}}$ for some $T \in \nat.$
\figf{fig-real_threat} illustrates such a datum for $T = 4.$
We assume the availability of a ``real-world'' dataset $\dataset = 
\{\datum^i\}_{i=1}^{\nData},$ where $\nData$ is small relative to the
typical dataset sizes used in DL.

The problem of interest is then formulated as follows:
\begin{prob}
	Given a training dataset $\dataset$ containing $\nData$ data points, 
	synthesize a new dataset $\datasetGen = \{\datum^j\}_{j=1}^{\nGen}$ such that
	$\datasetGen$ is statistically similar to $\dataset.$
\end{prob}
Implicit in this problem statement is the desire that the synthesis of samples 
in $\datasetGen$ should be computationally efficient, so that $\nGen$ can be 
made as large as needed.

\section{Generative AI Models}
\label{NN_architecture}

The proposed GAIM architecture to solve Problem~1 is a machine learning architecture
called the \mydef{split variational recurrent neural network (S-VRNN)},
which is based on the idea of a VRNN.
A VRNN merges the capabilities of recurrent neural networks (RNN) and variational 
autoencoders (VAE) to learn temporal dependencies in sequential data. 
In what follows, we describe the proposed S-VRNN based on brief descriptions 
of the VAE, RNN, and VRNN.

We assume that the reader is familiar with the idea of 
developing artificial neural networks as universal function approximators.
Briefly, a single-layer neural network (NN) may be considered to be a nonlinear 
function of the form $f(\datum; \nnParam) = \sigma(\nnParamWt^\intercal \datum + \nnParamBs),$ 
where $\vec{x}$ is the input, $\nnParam = (\nnParamWt, \nnParamBs)$ 
are parameters consisting of weights $\nnParamWt$ and biases $\nnParamBs,$ 
and $\sigma$ is a nonlinear activation function such as the sigmoid function.
The neural network \mydef{learns} or \mydef{is trained} over a dataset of input-output pairs
$\{(\datum^i, \yOutput^i)\}_{i=1}^{\nData}.$ Training is accomplished by finding parameters
$\nnParam^*$ that minimize a \mydef{loss function} $\loss$:
\begin{align*}
	\nnParam^* \defeq \arg \min_{\nnParam} \loss (\datum, \yOutput, \nnParam).
\end{align*}
The exact form of the loss function depends on the application. A common example
is the mean square loss function $\ell (\datum, \yOutput, \nnParam) \defeq \frac{1}{\nData} 
\sum_{i=1}^{\nData} \| \yOutput^i - f(\datum^i; \nnParam) \|^2. $

\subsection{Variational Autoencoders}
\label{ssec-vae}

A \mydef{variational autoencoder} (VAE) consists of two NNs called the
\mydef{encoder} $\encoder$ and \mydef{decoder} $\decoder$, respectively. The output space
of the encoder, which is also the input space of the decoder, is called the 
\mydef{latent space}, denoted $\latentSpace.$ The input space of the encoder,
which is also the output space of the decoder is the same as that of the data,
i.e., $\real[\nGridPts \times \nTime]$ in the present context.
To synthesize the desired dataset $\datasetGen,$
the decoder maps samples drawn from a standard normal distribution over 
the latent space $\latentSpace$ to its output space $\real[\nGridPts \times \nTime].$
The encoder learns a mapping from points 
$\datum \in \dataset$ to distributions in the latent space such that the distribution 
of $\latentVector \sim \encoder(\datum)$ conditioned on $\datum$
is approximately a standard normal distribution,
in the sense of low  Kullback-Leibler (KL) divergence.

More precisely, let $\nnParamEnc, \nnParamDec$ be
the parameters of the encoder and decoder NNs, respectively.
We denote by $p_{\nnParamDec}(\datum | \latentVector )$ the \mydef{likelihood}, 
i.e., the conditional distribution of the decoder's outputs $\datum$ 
given samples $\latentVector$ from the latent space. The objective of
statistical similarity between $\dataset$ and $\datasetGen,$ decoder
parameters are sought to maximize the log-likelihood.
Next, we denote by $q_{\nnParamEnc}(\latentVector | \datum )$ 
the conditional distribution of $\latentVector$ given $\datum.$
We can formulate this distribution as a normal distribution,
i.e., $q_{\nnParamEnc}(\latentVector | \datum ) \sim 
\mathcal{N}( \mu(\datum; \nnParamEnc), \Sigma(\datum; \nnParamEnc)),$
where $\mu$ and $\Sigma$ are the mean and covariance to be learned by
the encoder during training. The encoder and decoder are 
trained simultaneously by minimizing the loss
\begin{align}
	\loss\msub{VAE}(\nnParamEnc, \nnParamDec) &\defeq 
	- \E{\log p_{\nnParamDec} (\datum | \latentVector)}{z \sim q_{\nnParamEnc}} \nonumber \\
	& +
	\klDiv{\vphantom{1^{2}} 
	\mathcal{N} (\mu(\datum; \nnParamEnc), \Sigma(\datum; \nnParamEnc))~||~ 
	\mathcal{N}(0, I) }.
	\label{eq-vae-loss}
\end{align}
The first term in $\loss\msub{VAE}$ is a \emph{reconstruction loss,} which penalizes
outputs statistically dissimilar from the training data. The second term
in $\loss\msub{VAE}$ is a \emph{similarity loss,} which penalizes the 
difference of the learned latent space distribution to the decoder's sampling distribution
(standard normal).

\subsection{Recurrent Neural Networks}

A recurrent neural network (RNN) is a dynamical system where the time-dependent
state variable $\rnnState_t$ and its temporal evolution are learned. RNNs are designed
for \emph{temporal sequences} of inputs and outputs $\datum_t, \yOutput_t$
for a finite set of indices $t \in \nat.$
An RNN may be considered as a composition of two layers: a recurrent layer
with parameters $\nnParamDec$ and an output layer with parameters 
$\nnParam\msub{p}.$ The recurrent layer is a
mapping of the form $\rnnState_{t} = f(\datum_{t-1}, \rnnState_{t-1}; \nnParamDec),$ 
which involves feedback to the layer from the previous time step. 
This mapping may be called the \mydef{recurrence mapping}, and is 
similar to the right hand side of a typical state-space dynamical system 
differential- or difference equation. 
The recurrent layer is repeated for as many time steps as the length of the
input sequence $\datum_t.$ The output layer is a mapping of the form
$\yOutput_t = f\msub{p}(\datum_{t}, \rnnState_{t}; \nnParam\msub{p}).$
The RNN may be trained by minimizing a loss function that penalizes differences
between its outputs and desired outputs in the training dataset.

\subsection{Variational Recurrent Neural Networks}
\label{ssec-vrnn}

A variational recurrent neural network (VRNN) may be considered as
a combination of a VAE and an RNN. Like a VAE, the VRNN has encoder
and decoder NNs, denoted as before by $\encoder$ and $\decoder,$
with parameters $\nnParamEnc, \nnParamDec.$ Like an RNN, the VRNN
maintains a state $\rnnState_t$ and its recurrence mapping, and 
all quantities, namely, the input $\datum_t,$ the state $\rnnState_t,$
and the latent vector $\latentVector \sim \encoder(\datum)$ are
indexed by time.

The VRNN encoder approximates a distribution
$q_\nnParamEnc(\latentVector_t \mid \datum_t, \rnnState_{t-1})$
conditioned not only over the input~$\datum_t$ 
but also over the {state} $\rnnState_{t-1}.$
As in the VAE, we can formulate this distribution as 
a normal distribution, i.e.,
$q_{\nnParamEnc} \sim 
\mathcal{N}( \mu_t(\datum_t, \rnnState_{t-1}; \nnParamEnc), 
\Sigma_t(\datum; \nnParamEnc))$ to be learned.
The VRNN decoder draws samples from a standard normal distribution
over the latent space.
The generated output~$\datum_t$ is conditioned on the latent variable $\latentVector_t$
and the hidden state $\rnnState_{t-1}$, for the likelihood 
$p_\nnParamDec(\datum_t \mid \latentVector_t, \rnnState_{t-1})$. 
Finally, the state recurrence mapping is $\rnnState_t = f(\rnnState_{t-1}, 
\datum_t, \latentVector_t; \nnParamDec)$.

The VRNN training process to learn the mapping $f_\nnParamDec$ and the distributions
$p_\nnParamDec, q_\nnParamEnc$ follows by minimizing the loss:
\begin{align}
	\hspace{-1.5ex}
	\loss\msub{VRNN}(\nnParamEnc, \nnParamDec) &\defeq 
	- \E{ p_\nnParamDec(\datum_t \mid \latentVector_t, \rnnState_{t-1}) }{z \sim q_{\nnParamEnc}} 
	\nonumber  \\
	& + \klDiv{ q_{\phi}(\latentVector_t \mid \datum_t, \rnnState_{t-1})		
		\parallel \mathcal{N}(0, I)} .
	\label{eq-vrnn-loss_og}
\end{align}
This loss function is similar to $\loss\msub{VAE}$ in \eqn{eq-vae-loss}.
We approximate the reconstruction loss by a mean squared error.
For the given training dataset $\dataset,$ the loss
$\loss\msub{VRNN}(\nnParamEnc, \nnParamDec)$ is then calculated as
\begin{align}
	\hspace{-3ex}
	\loss\msub{VRNN}(\nnParamEnc, \nnParamDec) &=
	\mathbb{E}_{\datum \in \dataset}  \left[ \vphantom{1^{2^2}}
	\| \datum_{\leq t} - \decoder(\encoder(\datum_t)) \|^2 ~+ \right.  \nonumber \\%
	& \hspace{8ex} \left. \vphantom{1^{2^2}} \klDiv{\mathcal{N}( \mu_t, \Sigma_t) 
		\parallel \mathcal{N}(0, I)}
	\right].
	\label{eq-vrnn-loss}
\end{align}

\subsection{Split-Variational Recurrent Neural Networks}
\label{ssec-splitvrnn}

Notice that the training processes of the two GAIMs described so far, namely,
the VAE and the VRNN, are purely data-driven in that the dynamical system
equation \eqnnt{eq-dynamics} that underlies the training data is never used.
The main innovation of the proposed GAIM is 
the incorporation of the system dynamical model \eqnnt{eq-dynamics} in the 
training process. We will demonstrate in \scn{sec-results} that this
approach provides a solution to Problem~1 in situations where training 
data are scarce.

To this end, we augment the training dataset $\dataset$ with an additional synthetic
dataset $\datasetSup,$ which we refer to as the \mydef{support} dataset.
To synthesize data in the $\datasetSup,$ we first integrate
the dynamical equation \eqnnt{eq-dynamics} from a randomly chosen initial 
condition and \emph{with zero process noise}. The resulting trajectory $\threatState(t)$
is then transformed to an observation via \eqns{eq-threat}{eq-observation}
\emph{with zero measurement noise}. Due to this method of synthesis,
the support dataset $\datasetSup$ contains data points that are  \emph{noiseless}
but \emph{similar} to those in~$\dataset.$ Crucially, we may synthesize
an abundant number of data points in~$\datasetSup$ to augment $\dataset,$
which may be small.

The proposed innovation lies in the formulation and training of a new
GAIM that leverages $\datasetSup$ to make up for the lack of a large corpus of
real-world data in $\dataset.$
The proposed GAIM is similar to a VRNN. The difference is that 
we introduce two latent subspaces $\kappa_1$ and $\kappa_2$. 
These subspaces serve to disentangle the features common to 
the datasets $\datasetSup$ and $\dataset$ from the features unique
to each of them. The resulting model, which we refer to as the \mydef{split VRNN (S-VRNN)}
is designed to effectively leverage this structure in the latent space.

The latent subspace $\kappa_1$ is captures the latent-space representations of the 
real-world (noisy) training dataset, $\dataset$, whereas $\kappa_2$ is a shared 
latent-space representation of the augmented dataset $\dataset \cup 
\datasetSup.$ This structure allows $\kappa_1$ to 
learn the noise-related features specific to the original dataset $\dataset,$ 
identically as in the VRNN.
The subspace $\kappa_2$ allows the S-VRNN to learn common 
or transferable features across $\datasetSup$ and $\dataset$. 
Considering that $\datasetSup$ can be large, the shared latent 
suspace $\kappa_2$ learns generalizable features that support 
$\dataset$ as well, allowing the model to leverage 
patterns learned from $\datasetSup$ to improve representations for $\dataset$.

\begin{figure}
	\centering
	\includegraphics[width=0.8\columnwidth]{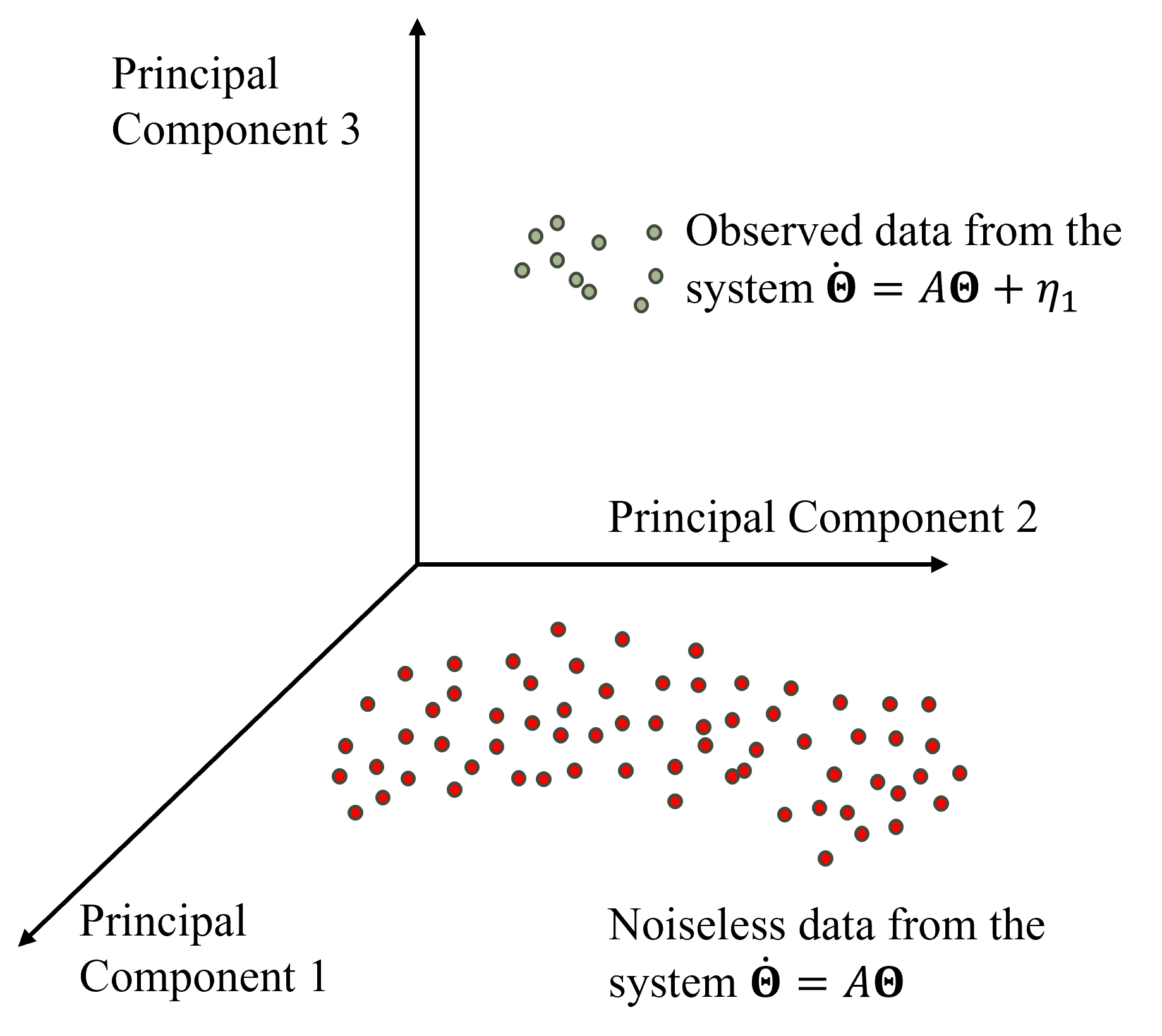}
	\caption{Illustration of S-VRNN training data.
	The S-VRNN architecture exploits the idea that, by definition, 
	the support dataset (red dots) lies in a sub-manifold of 
	the manifold formed by the dataset $\dataset$ (gray dots).}
	\label{fig-splitvrnn-graph}
\end{figure}

\figf{fig-splitvrnn-graph} illustrates the motivation behind the proposed S-VRNN
approach. A small number of data points in $\dataset,$ indicated in gray color,
belong to some manifold. A significantly larger number of data points in $\datasetSup,$
indicated in red, belong to a sub-manifold. Informally, the S-VRNN encodes the gray-colored
points in the latent subspace $\kappa_1,$ and \emph{all} points (gray and red)
in the subspace $\kappa_{2}.$

Informally, the latent variables in $\kappa_{2}$ are an approximation to 
the ``correct'' latent variables in $\kappa_1.$ Because $\datasetSup$ is
abundant, learning the S-VRNN parameters $\nnParamEnc, \nnParamDec$ using
$\datasetSup$ is not only easier but also provides an initial guess
for the ``correct'' values of parameters $\nnParamEnc, \nnParamDec.$

The S-VRNN encoder learns approximate distributions of each latent subspace
conditioned on the inputs $\datum_t$ and the hidden state $\rnnState_{t-1}$, denoted as: 
\begin{equation} 
	\begin{array}{ll}
		\kappa_1 \sim q^1_\nnParamEnc(\latentVector_t \mid \datum_t, 
		\rnnState_{t-1}), & \datum_t \in \dataset, \\ 
		\kappa_2 \sim q^2_\nnParamEnc(\latentVector_t \mid \datum_t, 
		\rnnState_{t-1}), & \datum_t \in \dataset
		\cup 
		\datasetSup
	\end{array} 
\end{equation}
We formulate these distributions to be normal, namely:
\begin{align*}
	q^1_{\nnParamEnc} &\sim 
	\mathcal{N}( \mu^1_t(\datum_t, \rnnState_{t-1}; \nnParamEnc), \Sigma^1_t(\datum_t, 
	\rnnState_{t-1}; \nnParamEnc)), \mbox{ and } \\
	q^2_{\nnParamEnc} &\sim 
	\mathcal{N}( \mu^2_t(\datum_t, \rnnState_{t-1}; \nnParamEnc), \Sigma^2_t(\datum_t, 
	\rnnState_{t-1}; \nnParamEnc)).
\end{align*}

The S-VRNN decoder samples from standard normal distributions over each
latent subspace.
By sampling from both latent subspaces, the decoder obtains information 
from the latent space unique  to $\kappa_1$, as well as from the shared space $\kappa_2$. 
This enables the generated data to 
contain features from the noiseless (dynamics-driven) and noisy (data-driven) data.
As before, the likelihood distribution over the decoder's output is 
$p_\nnParamDec(\datum_t \mid \latentVector_t, \rnnState_{t-1}).$
As in the VRNN, the S-VRNN also learns a recurrence mapping
$f(\rnnState_{t-1}, \datum_t, \latentVector_t; \nnParamDec)$.

We train the S-VRNN by minimizing the following loss:
\begin{align}
	\loss\msub{S-VRNN}(\nnParamEnc,\nnParamDec) &\defeq 
	\mathbb{E}_{\datum \in \dataset \cup \datasetSup} \left[ \vphantom{1^{2^2}}
		\| \datum_{\leq t} - \decoder(\encoder(\datum_{\leq t})) \|^2 ~+ \right.
	 	\nonumber \\
	& \left. \klDiv{\mathcal{N}( \mu^1_t, \Sigma^1_t) \parallel \mathcal{N}(0, I)} 
		\cdot \indicator ~+ \right. \nonumber \\
	& \left. \vphantom{1^{2^2}}
	\klDiv{\mathcal{N}( \mu^2_t, \Sigma^2_t) \parallel \mathcal{N}(0, I)} \right].
	\label{eq-splitvae-loss}
\end{align}
Here, $\indicator$ is an indicator function defined as
\begin{equation}
	\indicator \defeq 
	\left\{ 
	\begin{array}{ll}
		0, & \datum \in \dataset, \\ 
		1, & \datum \in \datasetSup. 
	\end{array} 
	\right.
\end{equation}

The proposed approach allows the GAIM to generate new samples by 
leveraging both shared and distinct information across the two datasets, 
effectively utilizing the common features while also preserving the unique 
characteristics of~$\dataset.$ 
This strategy is especially beneficial in scenarios where~$\dataset$ is small,
but~$\datasetSup$ can be as large as needed. By isolating and emphasizing 
distinctive characteristics, we enhance the GAIM's ability to generalize 
from a relatively small number of real-world training examples.

\section{Results \& Discussion}
\label{sec-results}

We implemented the proposed S-VRNN GAIM using PyTorch \cite{pytorchref}, 
which is a Python-based software library for implementing various architectures NN. 
For this study, the dataset $\dataset$ was synthetically produced, while using
a real-world dataset is a goal for future work.

To produce $\dataset,$ we repeatedly solved using \matlab\ the system dynamical 
equation~\eqnnt{eq-dynamics} with a zero-mean Gaussian white noise process $\eta_1.$
The dimension of the threat coefficient vector was fixed at $\nParam = 4.$
For each solution, the spatial basis parameters $a_i$ and $\vec{b}_i,$ 
for each $i \in \intset{\nParam}$ and the initial conditions $\threatState(0)$ 
of the threat state vector were randomly chosen. The measurement noise $\eta_2$
was neglected. The observation vector was recorded over a grid of size $100 \times 100,$
i.e., $\nGridPts = 10^4.$ Each observation vector was recorded over $10$ time steps.

We produced a pool of 500 such data points, from which we selected a smaller subset
of $\nData = 25, 50,$ or $100$ points to produce $\dataset$ with $T = 4$ time steps.

We implemented and compared three GAIMs: the proposed S-VRNN,
the VRNN described in \scn{ssec-vrnn}, and a split-VAE (S-VAE),
which is a modification of the VAE described in \scn{ssec-vae}
using the split latent space idea described in \scn{ssec-splitvrnn}.
Note that the S-VRNN and S-VAE are GAIMs that learn from the data
as well the noiseless system dynamical equation, whereas the VRNN
learns from data, only. Hyperparameters such as the number of layers,
dimensions of each layer, etc., for each GAIM were established
after extensive numerical experimentation, and are described next.

\subsection{GAIM Architecure Details}

The VRNN architecture was implemented with a 2-layer encoder 
and a 2-layer decoder. \tbl{tbl-vrnn-layers} indicates the dimensions
of the input, output, and hidden layers (H$_i$). Each layer used
the $\tanh$ activation function.
Note that the encoder input and decoder output sizes
are $\nGridPts = 10^4.$ The latent space (encoder output and decoder input)
dimension was set to $16.$
\begin{table}
	\centering
	\caption{Layer dimensions of VRNN implementation.}
	\label{tbl-vrnn-layers}
	\begin{tabular}{l|lll}
		\toprule[1pt]
		NN Layers & Input & H$_1$  & Output \\
		\midrule[1pt]
		Encoder & $10^4$ & $40$ & $16$  \\
		Decoder& $16$ & $40$ & $10^4$   \\ 
		\bottomrule[1pt]
	\end{tabular}
\end{table}
The RNN sequence length was set to $4.$ Layer normalization was applied
to the hidden states after the RNN encoder and decoder, ensuring that the
outputs are normalized before proceeding to the next stages.

The S-VRNN architecture was implemented similar to the VRNN. 
The S-VRNN encoder and decoder each had four layers with dimensions
indicated in \tbl{tbl-svrnn-layers}. The S-VRNN latent space (encoder 
output and decoder input) dimension was set to $20$ for each
of the two  subspaces $\kappa_{1}$ and~$\kappa_{2}.$
\begin{table}
	\centering
	\caption{Layer dimensions of S-VRNN implementation.}
	\label{tbl-svrnn-layers}
	\begin{tabular}{l|lllll}
		\toprule
		& Input & H$_1$ & H$_2$ & H$_3$ &  Output \\
		\midrule
		Encoder & $10^4$ & $40$ & $80$ & $40$ & $20, 20$  \\
		Decoder & $20, 20$ & $40$ & $80$ & $40$ & $10^4$  \\ 
		\bottomrule
	\end{tabular}
\end{table}

\begin{table}
	\centering
	\caption{Layer dimensions of S-VAE implementation.}
	\label{tbl-svae-layers}
	\begin{tabular}{l|l}
		\toprule
		{Layer}            & {Output Dimensions}         \\ \midrule
		\multicolumn{2}{c}{{Encoder}} \\ \midrule
		Input                     & $(4, 100, 100)$                      \\
		Conv2d Layer 1            & $(16, 50, 50) $                      \\  
		Conv2d Layer 2            & $(32, 25, 25) $                      \\  
		Conv2d Layer 3            & $(64, 13, 13)$                       \\  
		Conv2d Layer 4            & $(128, 7, 7)$                        \\  
		Fully Connected Layer     & $(16)$                               \\ \midrule
		\multicolumn{2}{c}{{Decoder}} \\ \midrule
		Fully Connected Layer     & $(128, 7, 7) $                       \\  
		ConvTranspose2d Layer 1   & $(64, 14, 14) $                      \\  
		ConvTranspose2d Layer 2   & $(32, 28, 28)  $                     \\  
		ConvTranspose2d Layer 3   & $(16, 56, 56)  $                     \\  
		ConvTranspose2d Layer 4   & $(4, 100, 100) $                     \\  
		\bottomrule
	\end{tabular}
	\label{tab:vae_dimensions_simple}
\end{table}

The S-VAE was implemented using a convolutional neural network (CNN) architecture, 
with 4 channels to capture the threat field magnitude at specified time instances, 
to capture the temporal nature of the data. The training data were organized as a 
tensor of dimensions $[\nData \times T \times \sqrt{\nGridPts} \times \sqrt{\nGridPts}].$ 

The encoder includes 4 convolutional layers, while the decoder consists of one
fully connected layer followed by 4 transposed convolutional layers, with rectified 
linear unit (ReLU)  activation functions. 
\tbl{tbl-svae-layers} indicates the dimensions
of the input, output, and hidden layers.

For training each of these three GAIMs, 
a mini-batch size of $10$ and learning rate of $10^{-3}$ were chosen.

\subsection{Synthetic Data Generation Results}

To evaluate the performance of the three GAIMs, we visualize the 
training- and generated datasets by plotting the data points along the
axes with the three largest principal components. 
Additionally, we calculate the first four statistical moments 
of the training- and generated datasets for analyzing their 
statistical similarity.

\begin{figure}
	\centering
	\subfigure{\includegraphics[width=\columnwidth]{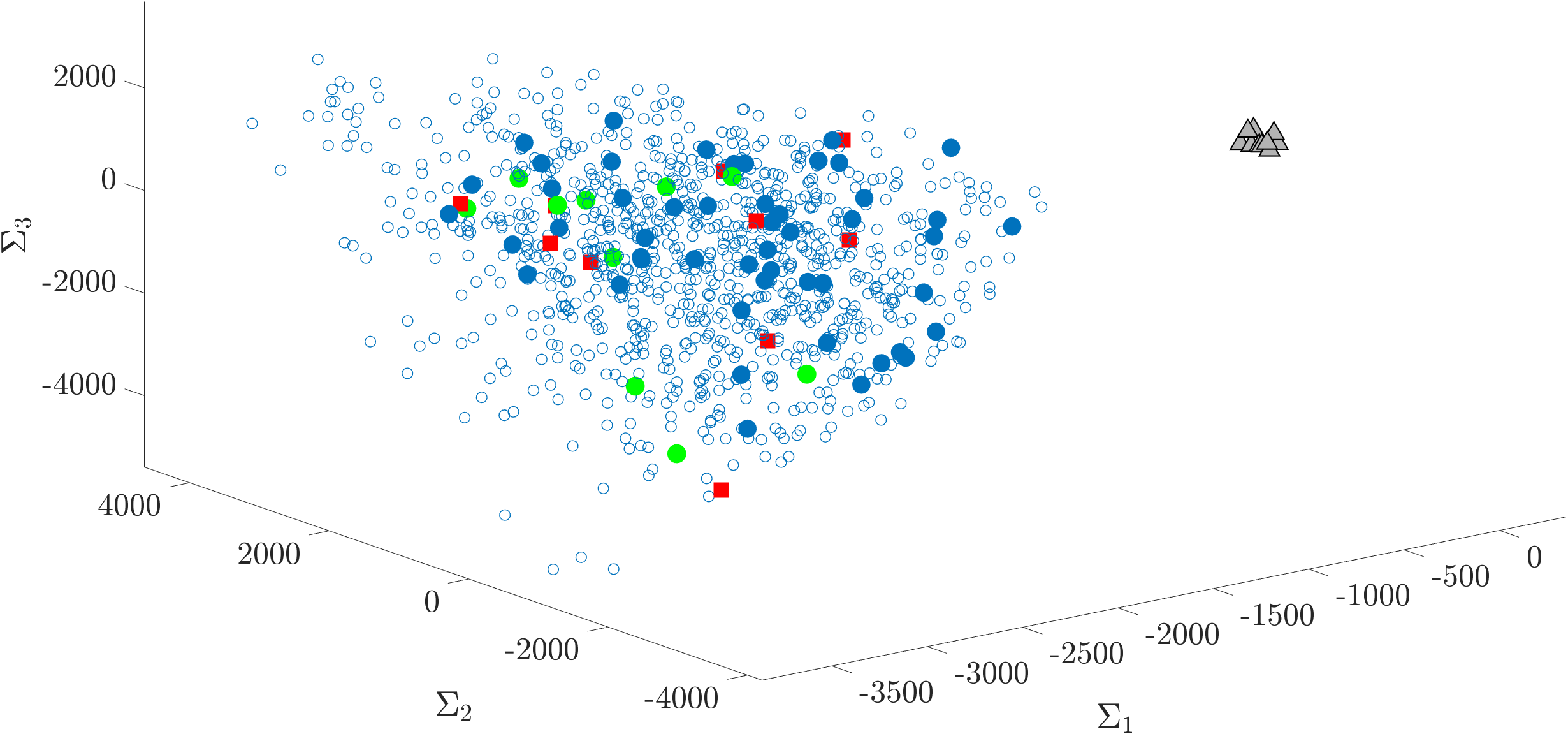}}
	\hspace{0.02\columnwidth}
	\subfigure{\includegraphics[width=\columnwidth]{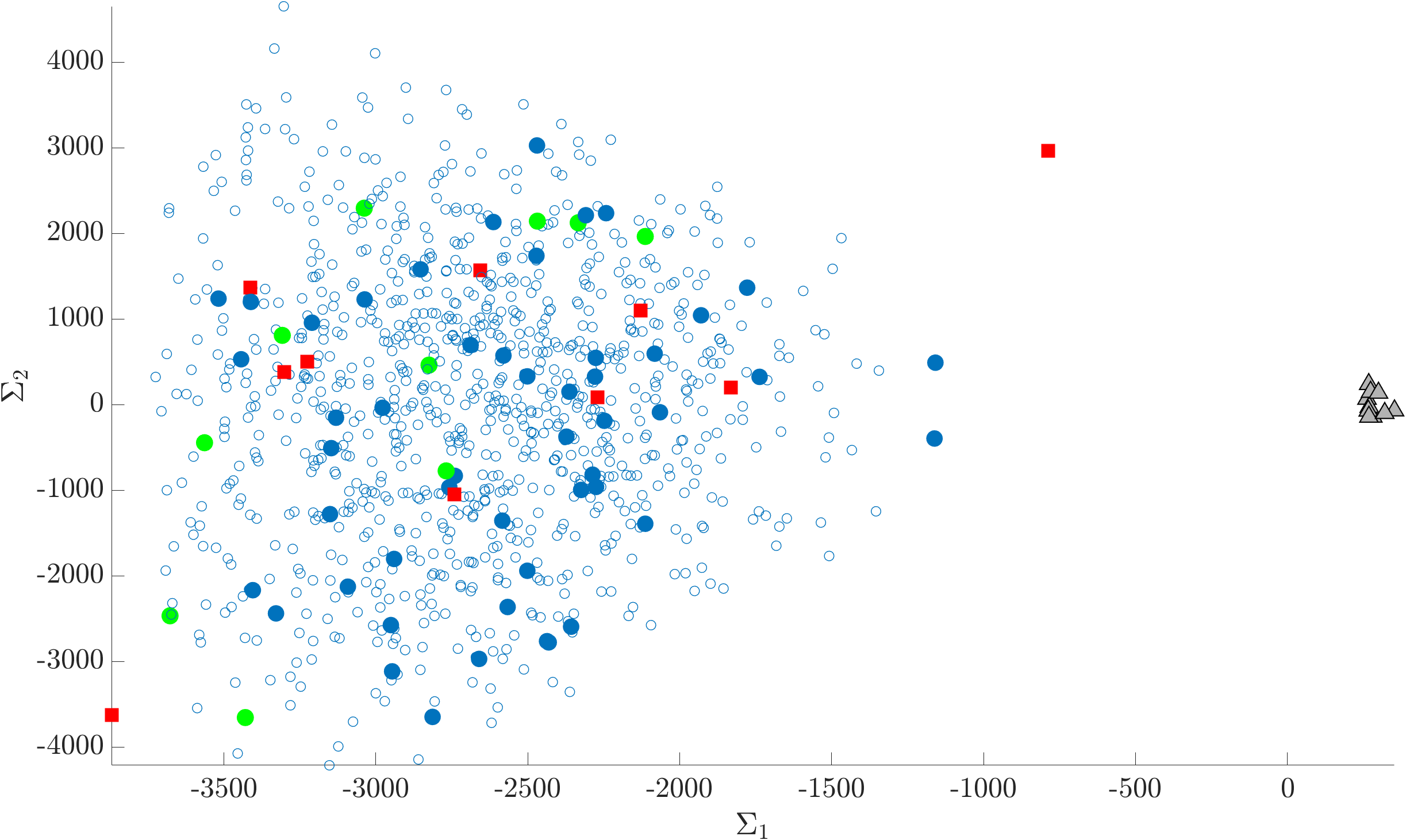}}
	\hspace{0.02\columnwidth}
	\subfigure{\includegraphics[width=\columnwidth]{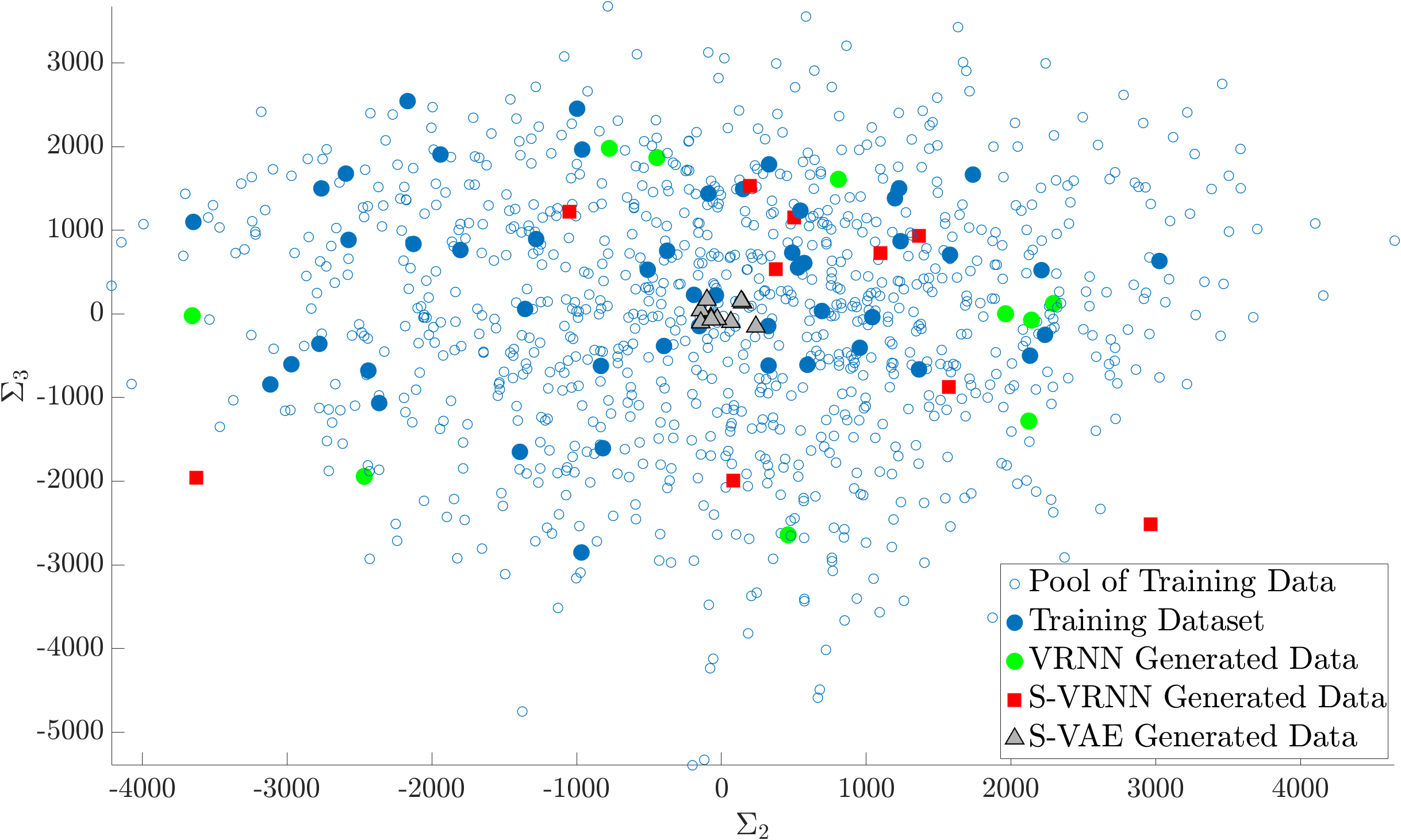}}
	\caption{{Visualization of the training- and generated data for $\nData = 50.$}}
	\label{fig-metric_50}
\end{figure}

\begin{figure}
	\centering
	\subfigure{\includegraphics[width=\columnwidth]{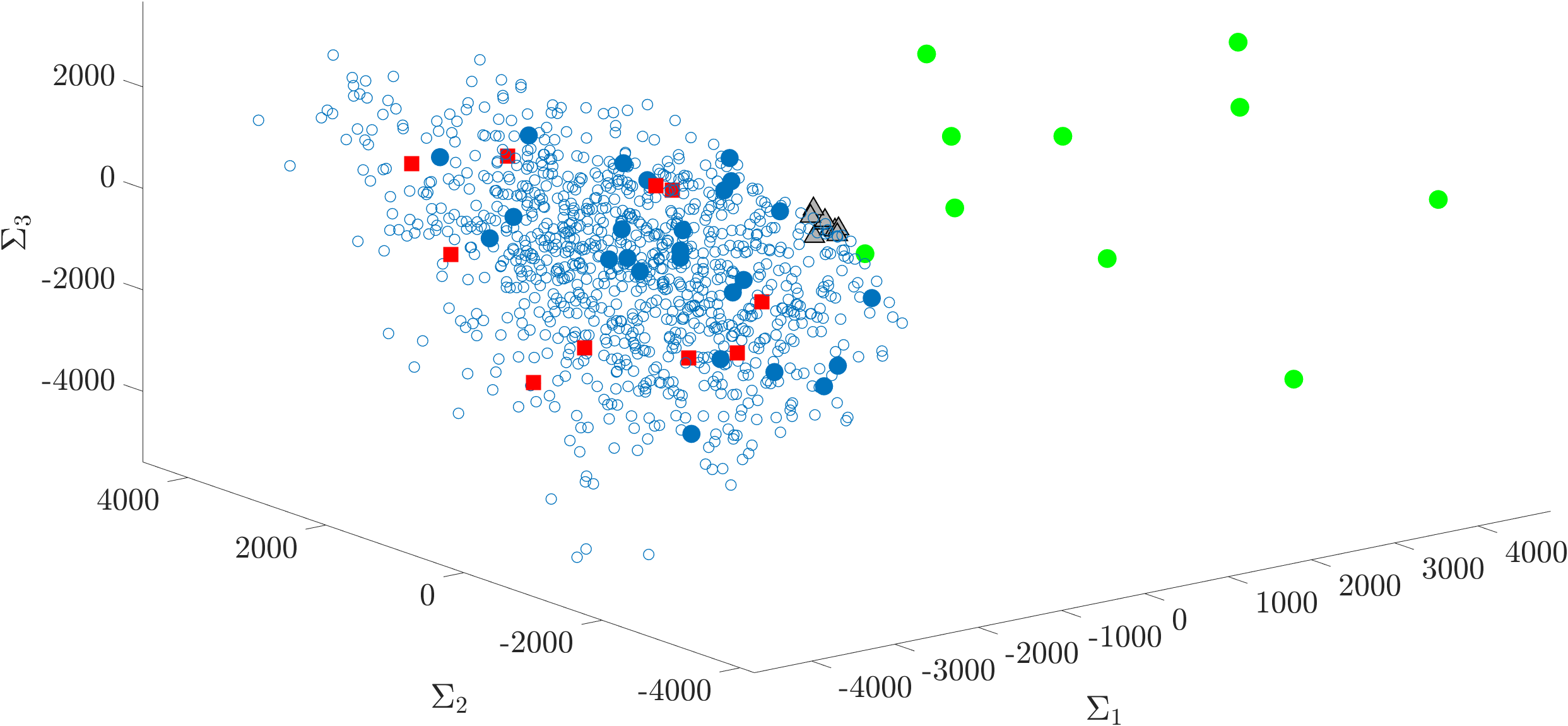}}
	\hspace{0.02\columnwidth}
	\subfigure{\includegraphics[width=\columnwidth]{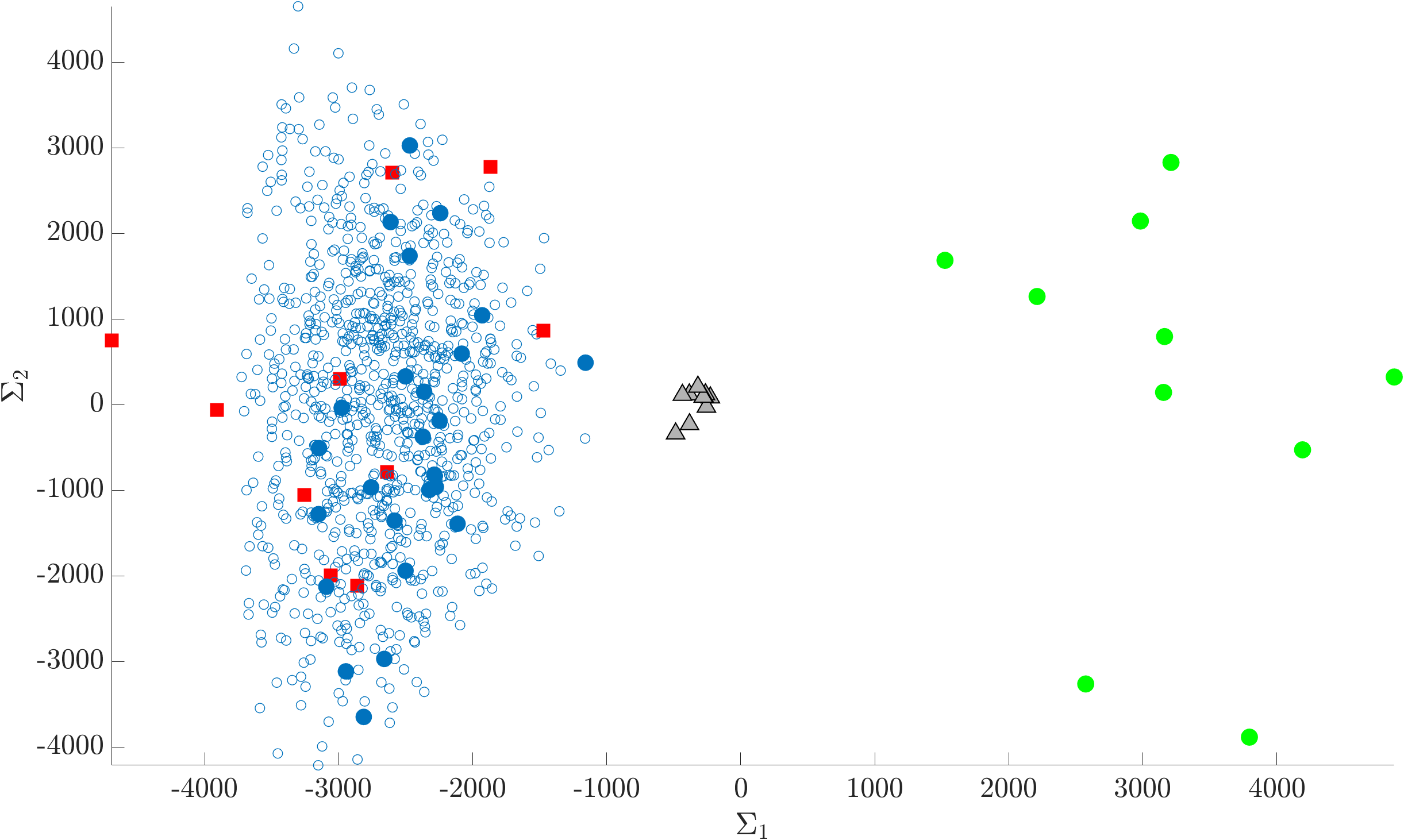}}
	\hspace{0.02\columnwidth}
	\subfigure{\includegraphics[width=\columnwidth]{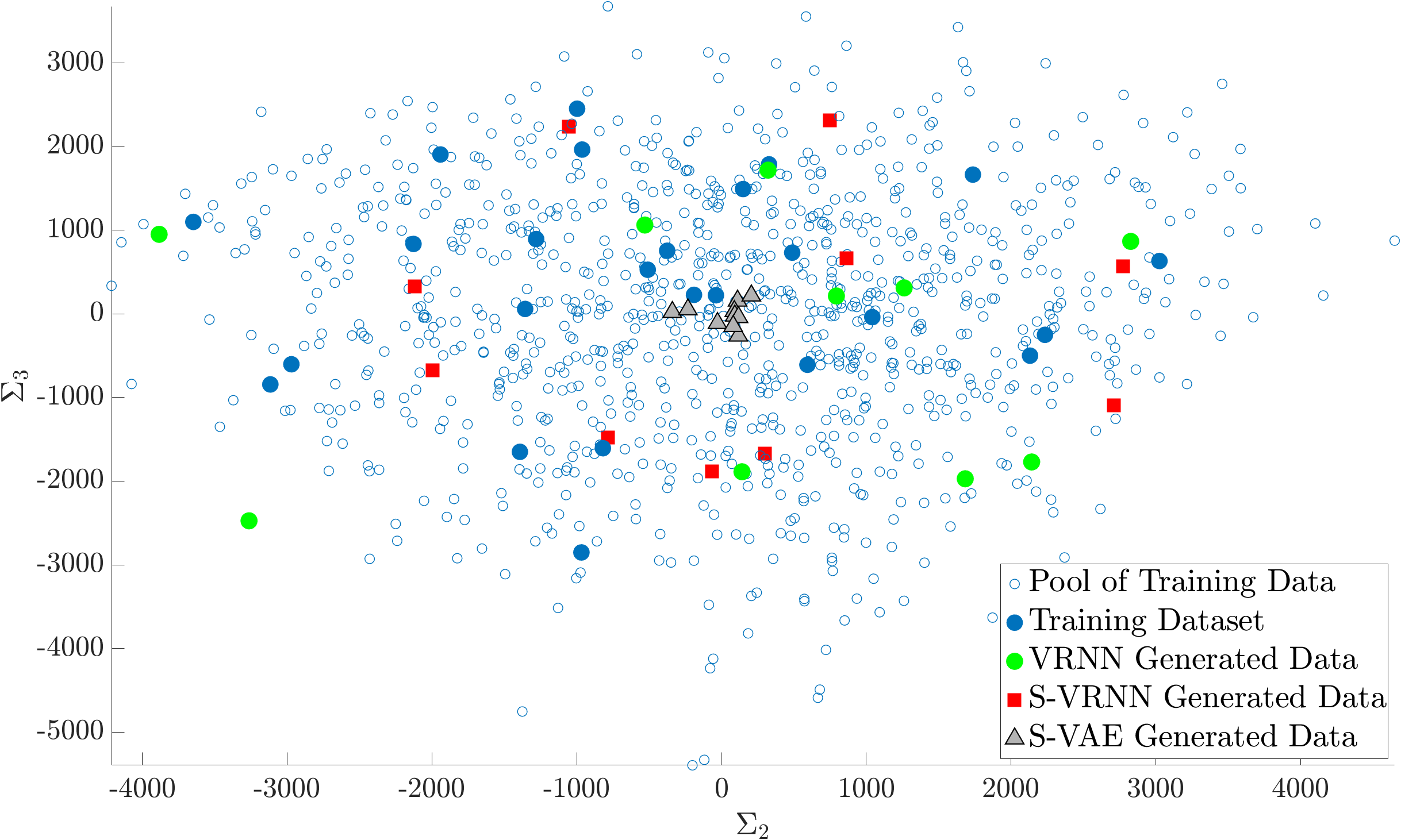}}
	\caption{{Visualization of the training- and generated data for $\nData = 25.$}}
	\label{fig-metric_25}
\end{figure}

First, consider the case with $\nData = 50.$ \figf{fig-metric_50}
visualizes the generated samples from the three GAIMs in comparison
to the pool of training data points (indicated by blue circles) and the 
dataset $\dataset$ (indicated by filled blue dots). Outputs from the S-VAE 
are indicated by black triangles, the VRNN by green dots, and the 
proposed S-VRNN by red squares.
Notice that the VRNN and S-VRNN output samples are close to, 
but not identical to, the points in the pool of training data.
By contrast, the S-VAE outputs perform poorly and are not close 
to the training data manifold. The three plots in \fig{fig-metric_50}
show the same 3D plot from different perspectives.
\tbl{tbl-moments-nd50} shows statistical moments along the
largest principal components up to four significant digits. 
Notice that the VRNN and S-VRNN
moments are similar to those of the training data pool, whereas the
S-VAE sample means and variances differ by one or more orders of magnitude.


\begin{table*}
	\centering
	\caption{Statistical moments of the data samples for $\nData = 50.$}
	\label{tbl-moments-nd50}
	\resizebox{\textwidth}{!}{%
			\begin{tabular}{lrrr|rrr|rrr|rrr}
					\toprule
					\textbf{} & \multicolumn{3}{c|}{\textbf{Mean$(\mu)$}} & 
					\multicolumn{3}{c|}{\textbf{Variance$(\sigma^2)$}} & 
					\multicolumn{3}{c|}{\textbf{Skewness$(\gamma)$}} & 
					\multicolumn{3}{c}{\textbf{Kurtosis$(\kappa)$}} \\
					\midrule
					\textbf{} & $\mu_{\Sigma_{1}}$ & $\mu_{\Sigma_{2}}$ & $\mu_{\Sigma_{3}}$ & 
					$\sigma^2_{\Sigma_{1}}$ & $\sigma^2_{\Sigma_{2}}$ & $\sigma^2_{\Sigma_{3}}$ & 
					$\gamma_{\Sigma_{1}}$ & $\gamma_{\Sigma_{2}}$ & $\gamma_{\Sigma_{3}}$ & 
					$\kappa_{\Sigma_{1}}$ & $\kappa_{\Sigma_{2}}$ & $\kappa_{\Sigma_{3}}$ \\
					\midrule
					Training data pool & $-2.681\tento{3}$ & $9.65$ & $-11.81$ & $0.2381\tento{6}$ & 
					$2.498\tento{6}$ & 
					$2.153\tento{6}$ & 
					$0.17$ & $-0.07$ & $-0.39$ & $2.62$ & $2.61$ & $3.05$ \\
					S-VAE generated data & $286.4$ & $3.06$ & $-2.95$ & $872.13$ & $18.15\tento{3}$ & 
					$13984.3$ 
					& $1.34$ & $0.50$ & $0.41$ & $3.51$ & $1.84$ & $1.60$ \\
					VRNN generated data & $-2.953\tento{3}$ & $245.0$ & $-37.44$ & $0.2924\tento{6}$ & 
					$4.302\tento{6}$ & 
					$2.490\tento{6}$ & $0.14$ & $-0.70$ & $-0.21$ & $1.71$ & $2.26$ & $1.97$ \\
					S-VRNN generated data & $-2.622\tento{3}$ & $348.4$ & $-122.8$ & $0.8210\tento{6}$ & 
					$3.091\tento{6}$ & $2.402\tento{6}$ & $0.61$ & $-0.96$ & $-0.49$ & $2.75$ & $3.93$ & 
					$1.54$ \\
					\bottomrule
				\end{tabular}
		}
\end{table*}

Next, consider the case with $\nData = 25,$ i.e., a smaller volume of
noisy training data compared to the previous case. 
\figf{fig-metric_25} is a visualization of the training- and generated
data for this case, similar to \fig{fig-metric_50}.
\tbl{tbl-moments-nd25} shows statistical moments along the
largest principal components up to four significant digits. 
These results indicate that the S-VAE, as in the previous case, performs
poorly in the sense that its generated outputs are highly dissimilar
compared to the pool of training data. 
Importantly, notice in \fig{fig-metric_25} and \tbl{tbl-moments-nd25}
that the \emph{purely data-driven VRNN also performs poorly.}
By contrast, the proposed S-VRNN continues to generate statistically
similar samples despite the smaller volume of training data owing 
to the incorporation of the underlying system dynamical equation
in its training.

\begin{figure}
	\centering
	\subfigure[$t =1$]{\includegraphics[width=0.46\columnwidth]{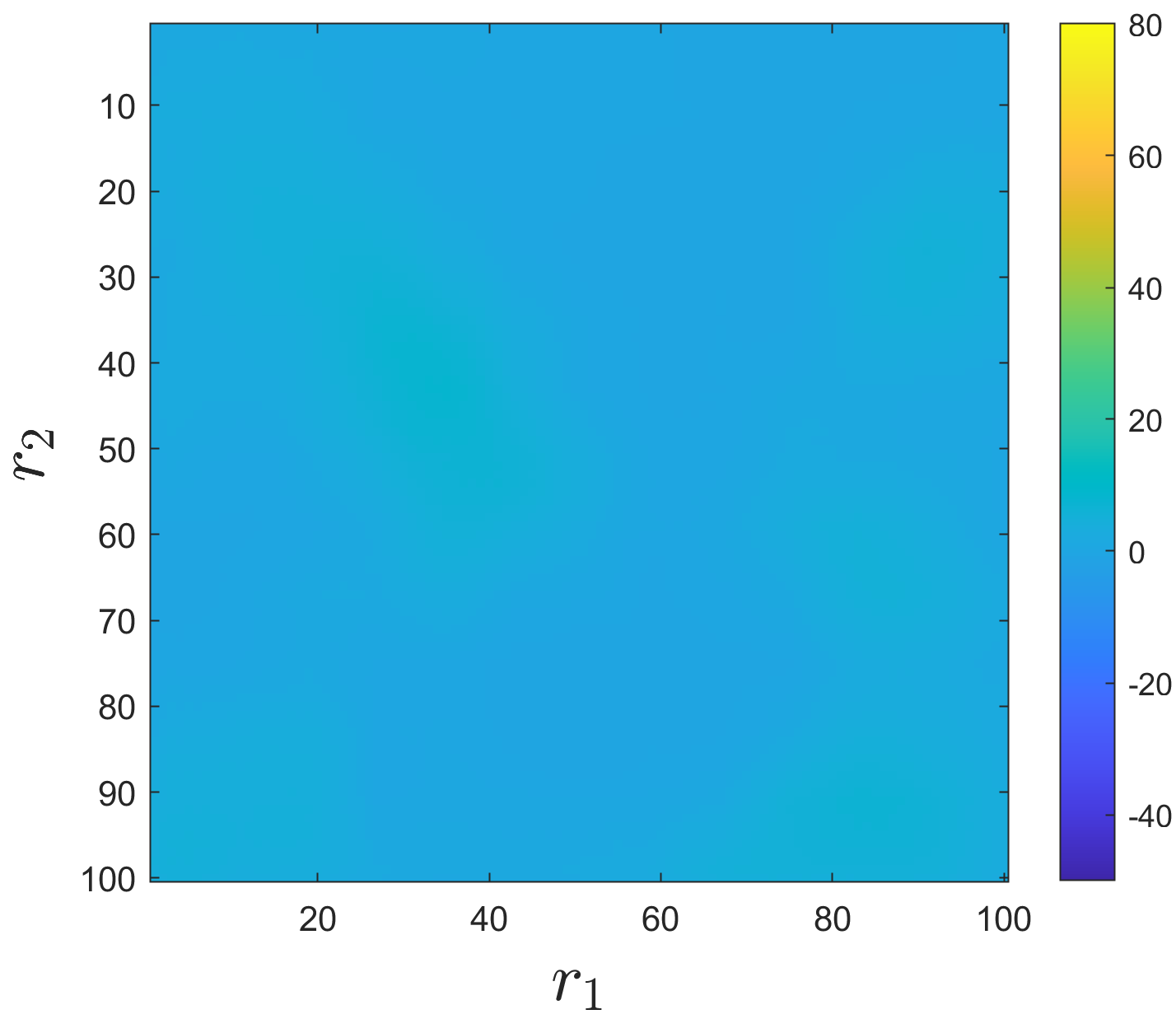}}
	\hspace{0.02\columnwidth}
	\subfigure[$t = 2$]{\includegraphics[width=0.46\columnwidth]{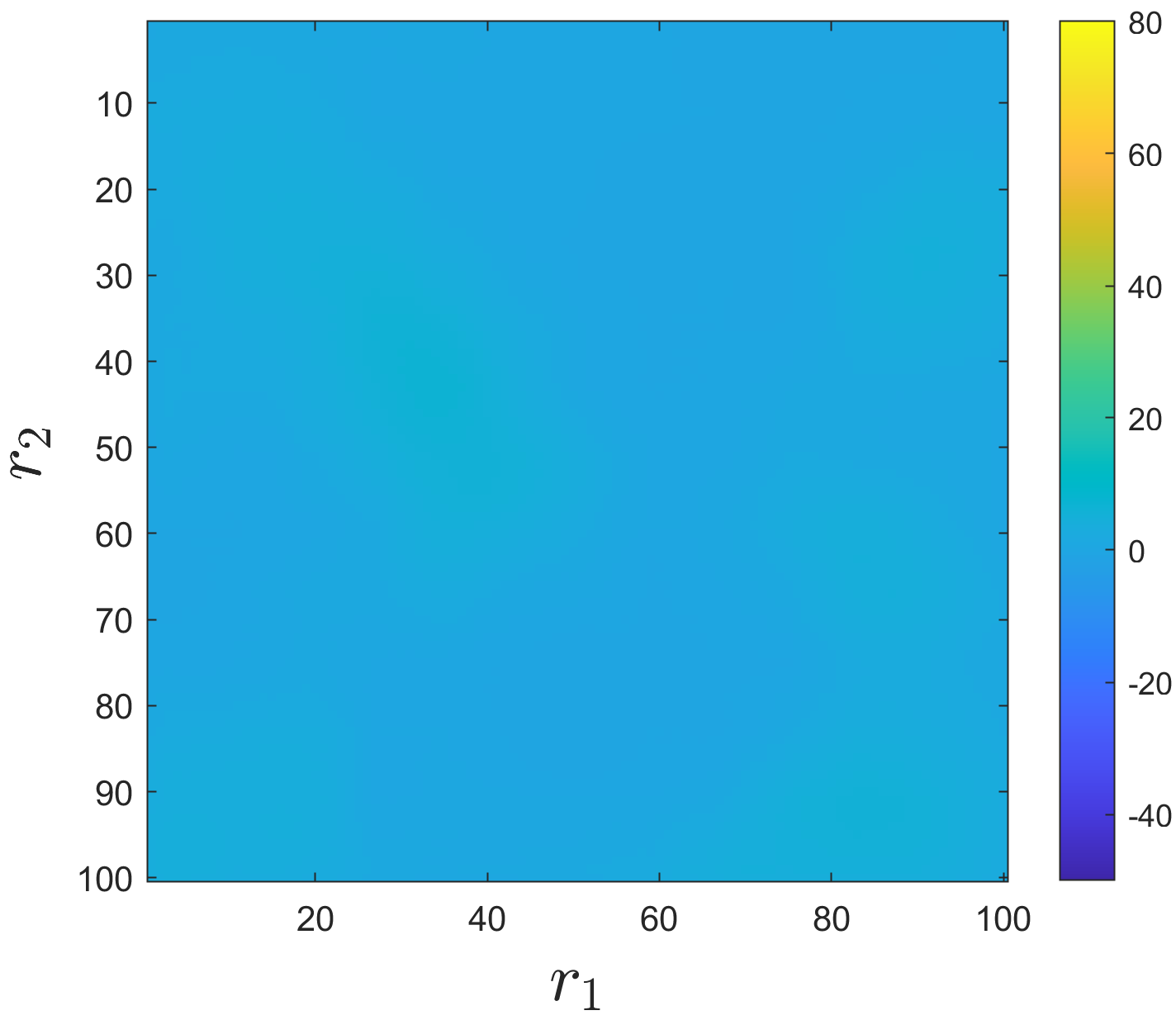}}
	\hspace{0.02\columnwidth}
	\subfigure[$t = 3$]{\includegraphics[width=0.46\columnwidth]{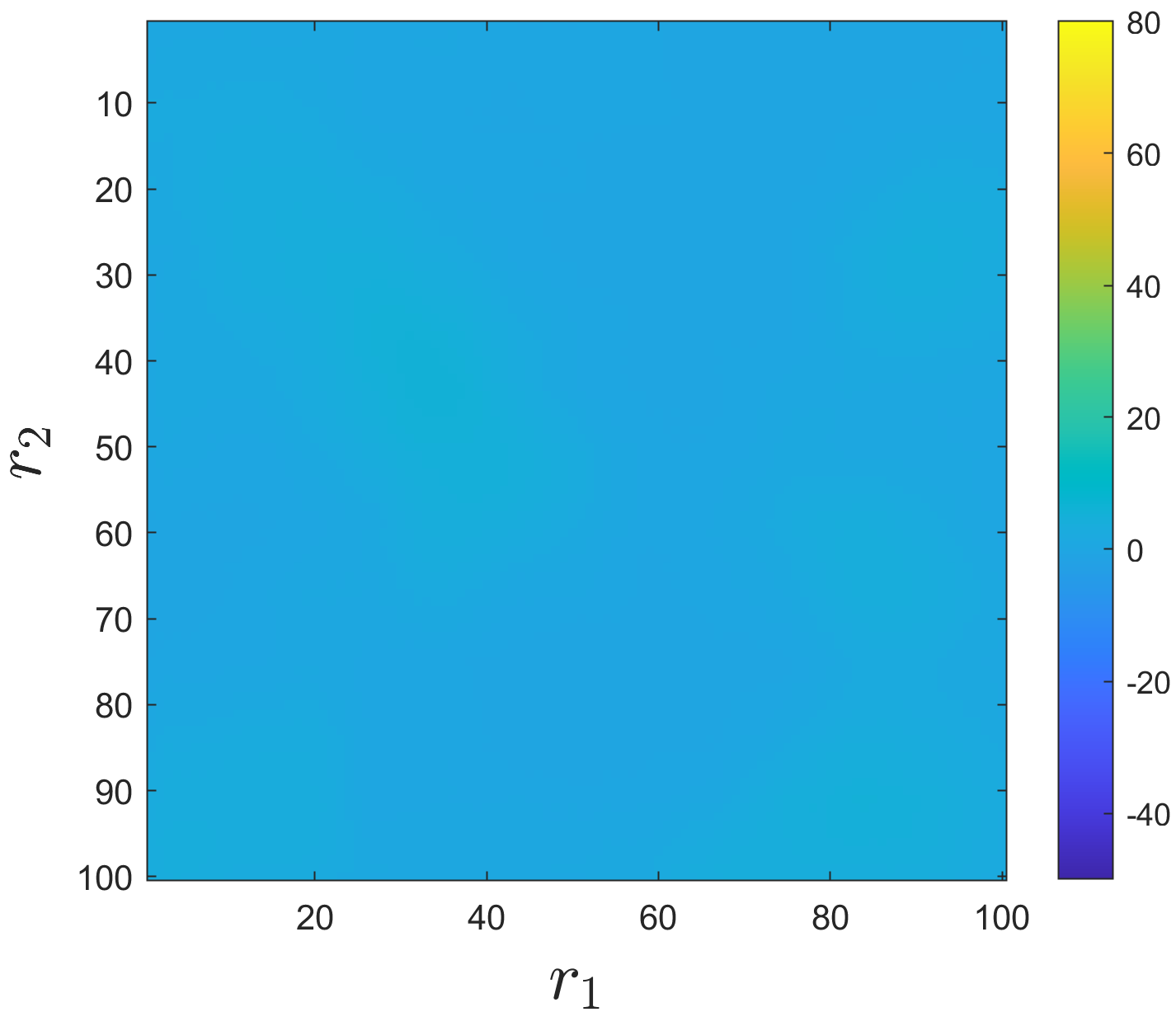}}
	\hspace{0.02\columnwidth}
	\subfigure[$t = 4$]{\includegraphics[width=0.46\columnwidth]{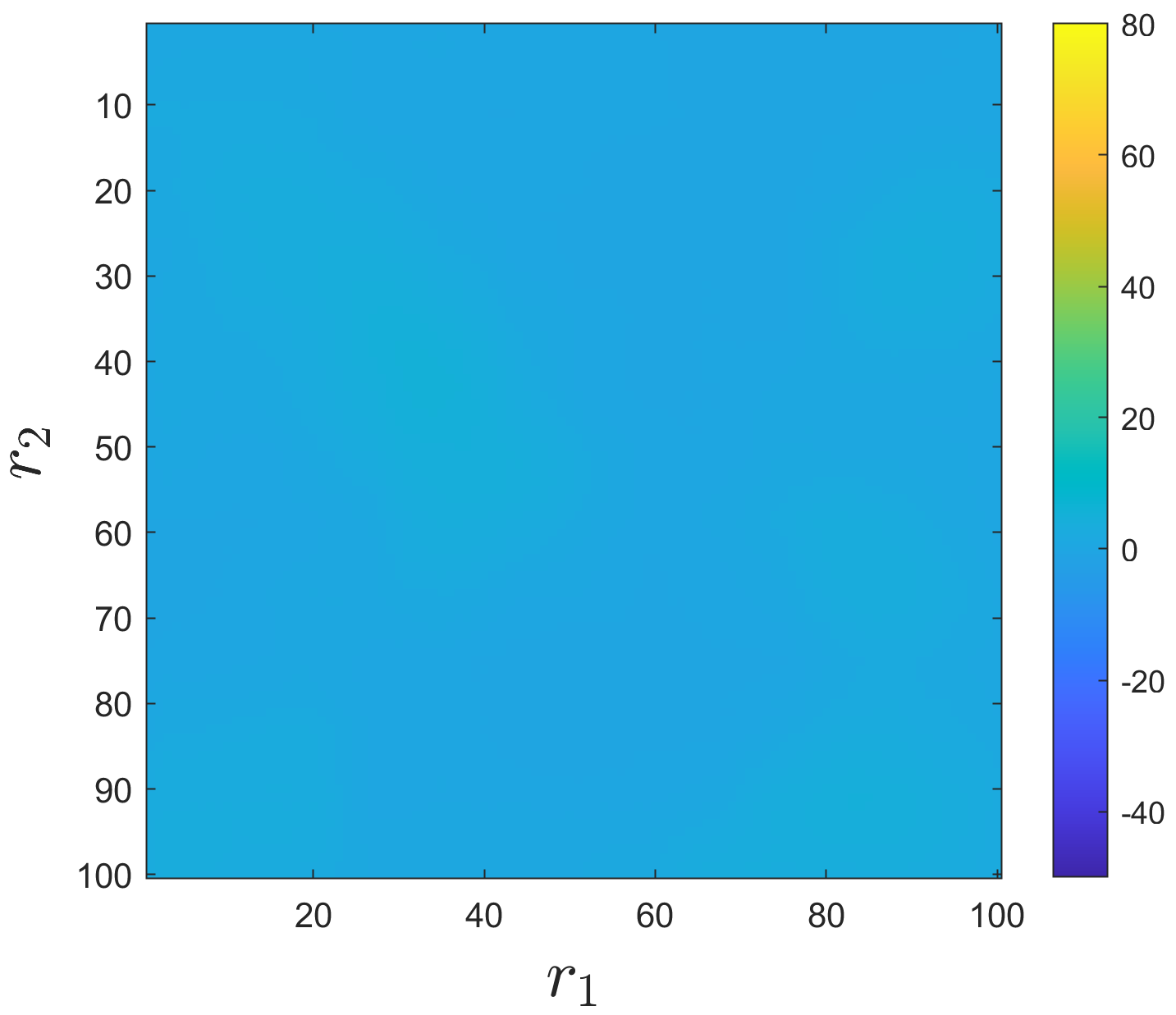}}
	\caption{S-VAE generated samples for $\nData = 25.$}
	
	\label{fig-gen_threat_svae}
\end{figure}
\begin{figure}
	\centering
	\subfigure[$t =1$]{\includegraphics[width=0.46\columnwidth]{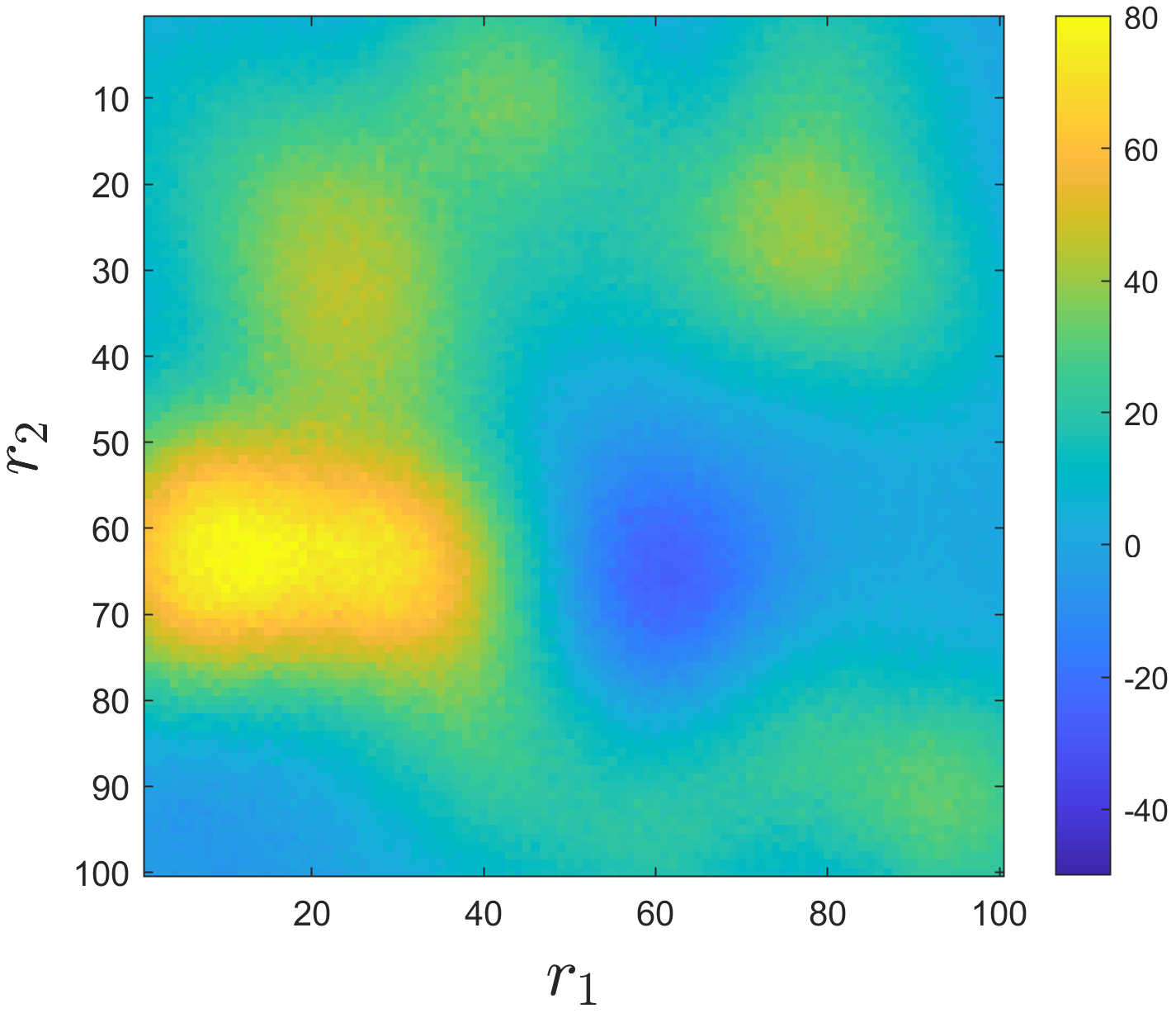}}
	\hspace{0.02\columnwidth}
	\subfigure[$t = 2$]{\includegraphics[width=0.46\columnwidth]{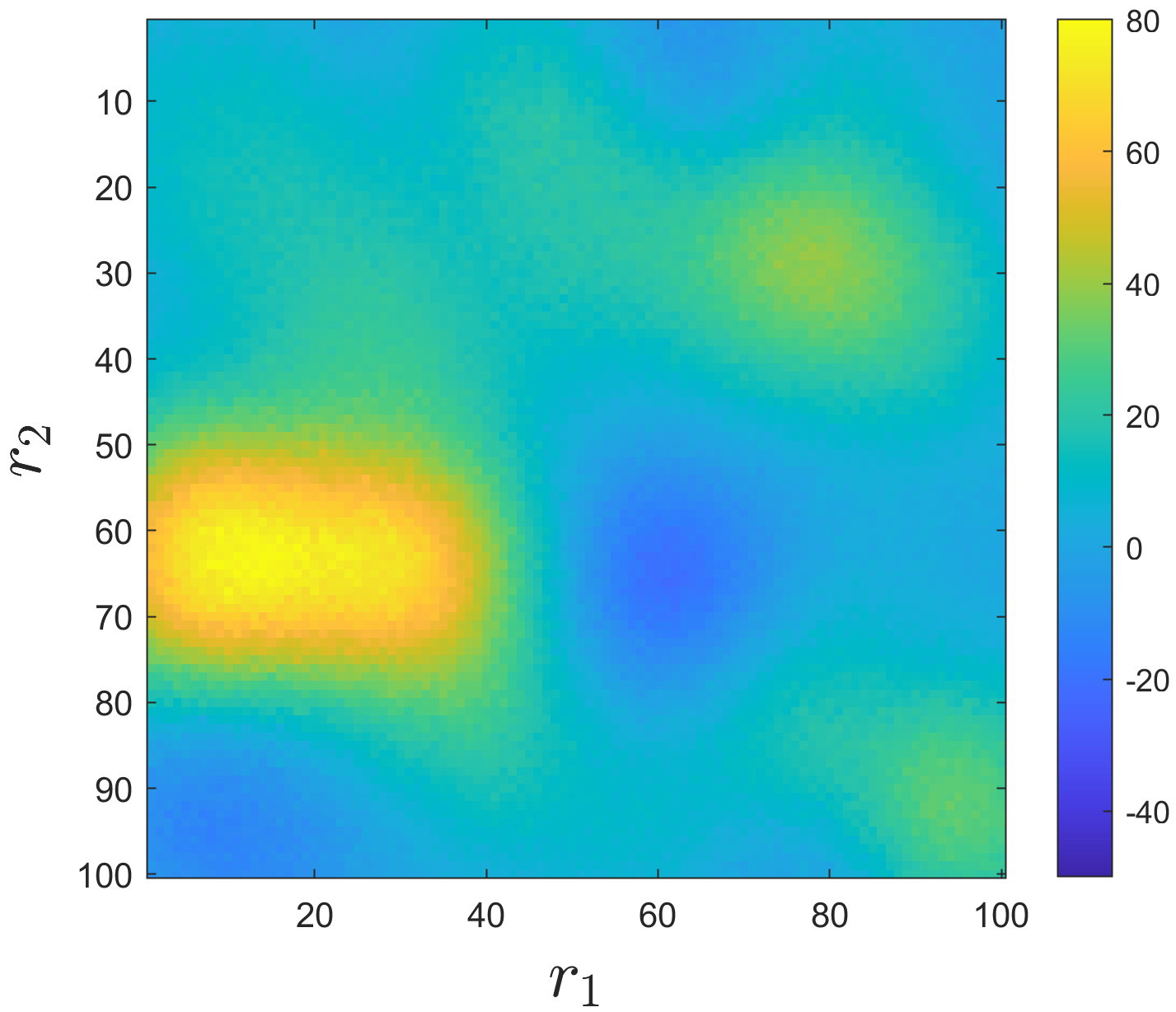}}
	\hspace{0.02\columnwidth}
	\subfigure[$t = 3$]{\includegraphics[width=0.46\columnwidth]{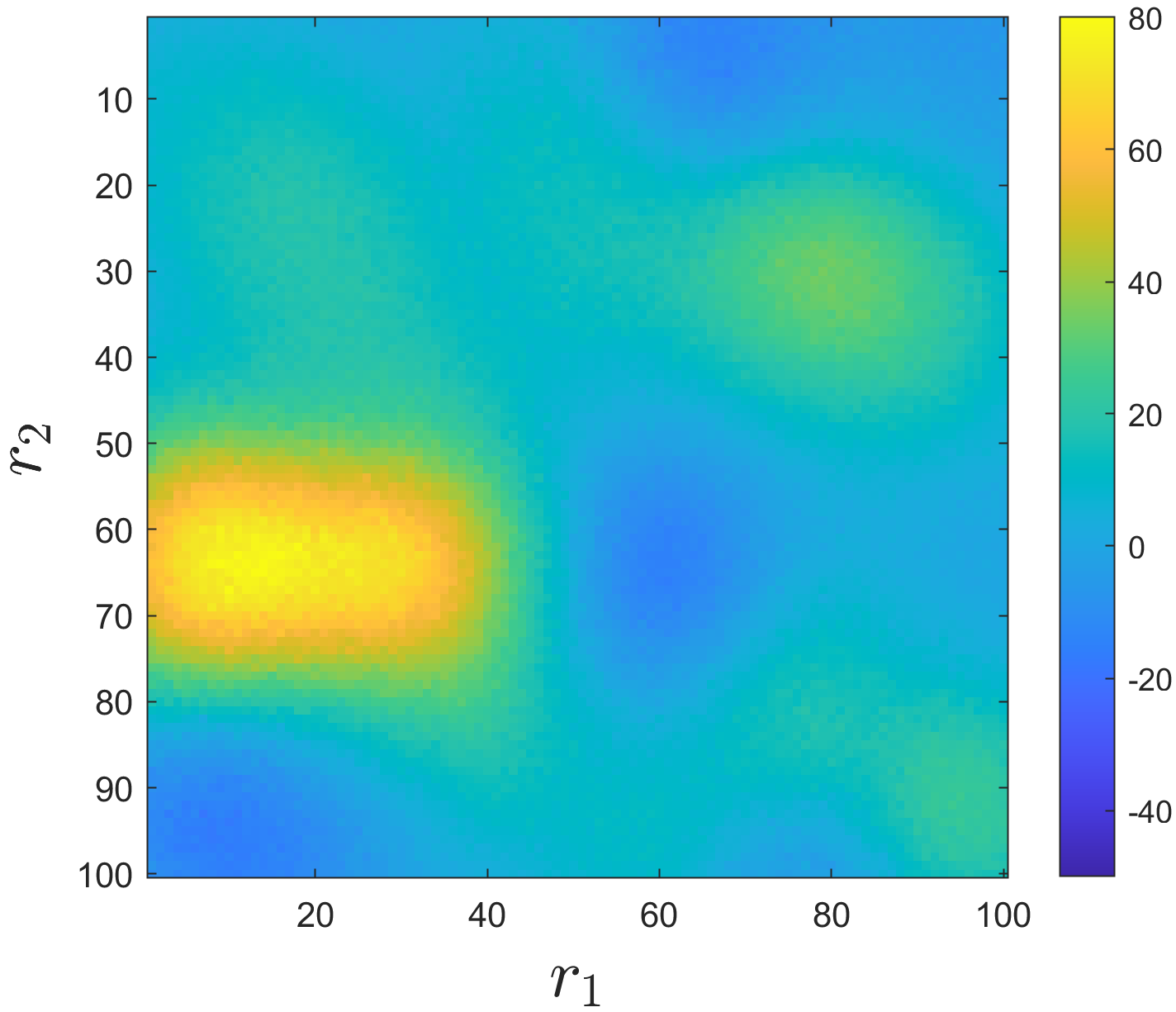}}
	\hspace{0.02\columnwidth}
	\subfigure[$t = 4$]{\includegraphics[width=0.46\columnwidth]{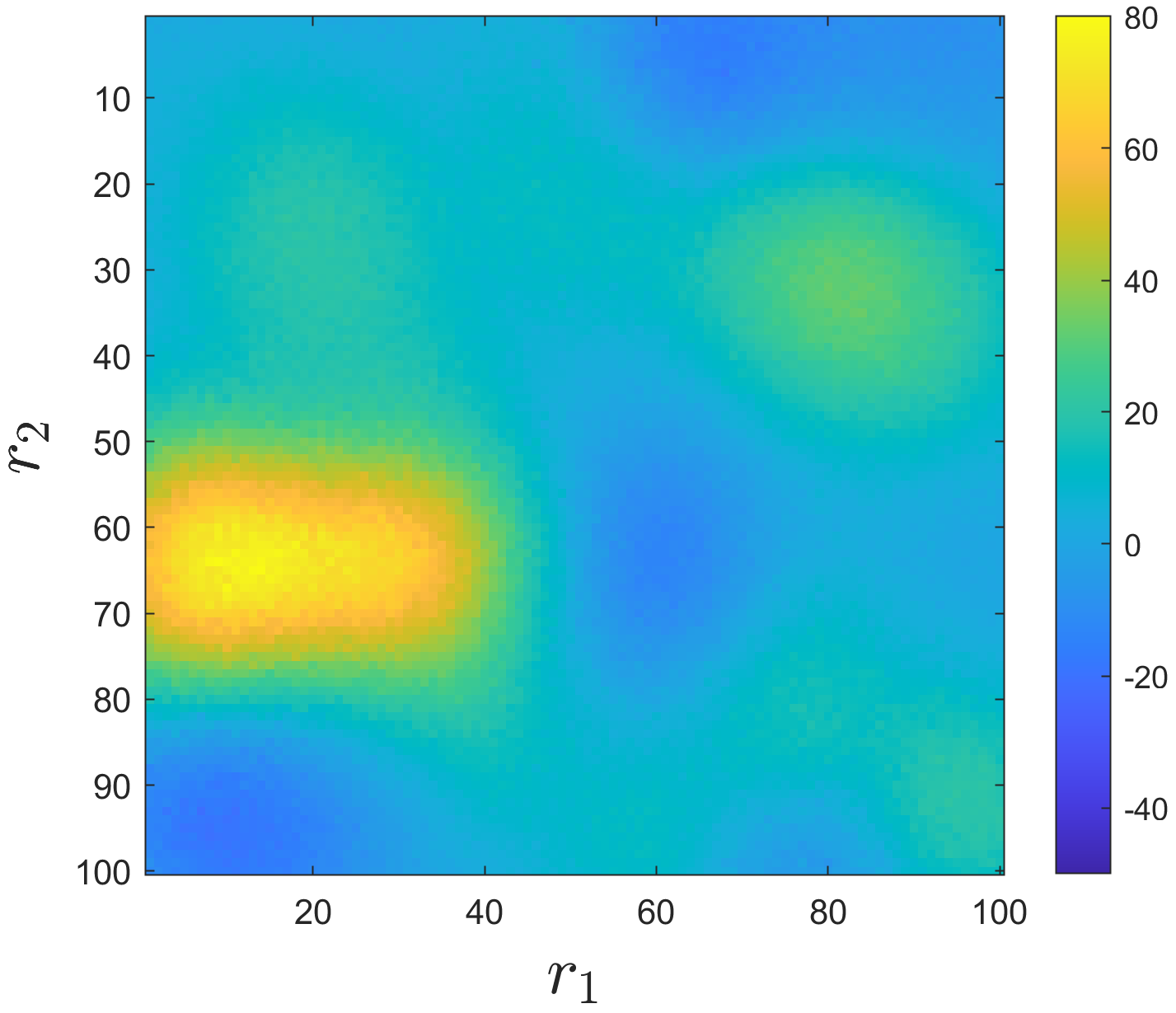}}
	\caption{VRNN generated samples for $\nData = 25.$}
	
	\label{fig-gen_threat_vrnn}
\end{figure}
\begin{figure}
	\centering
	\subfigure[$t =1$]{\includegraphics[width=0.46\columnwidth]{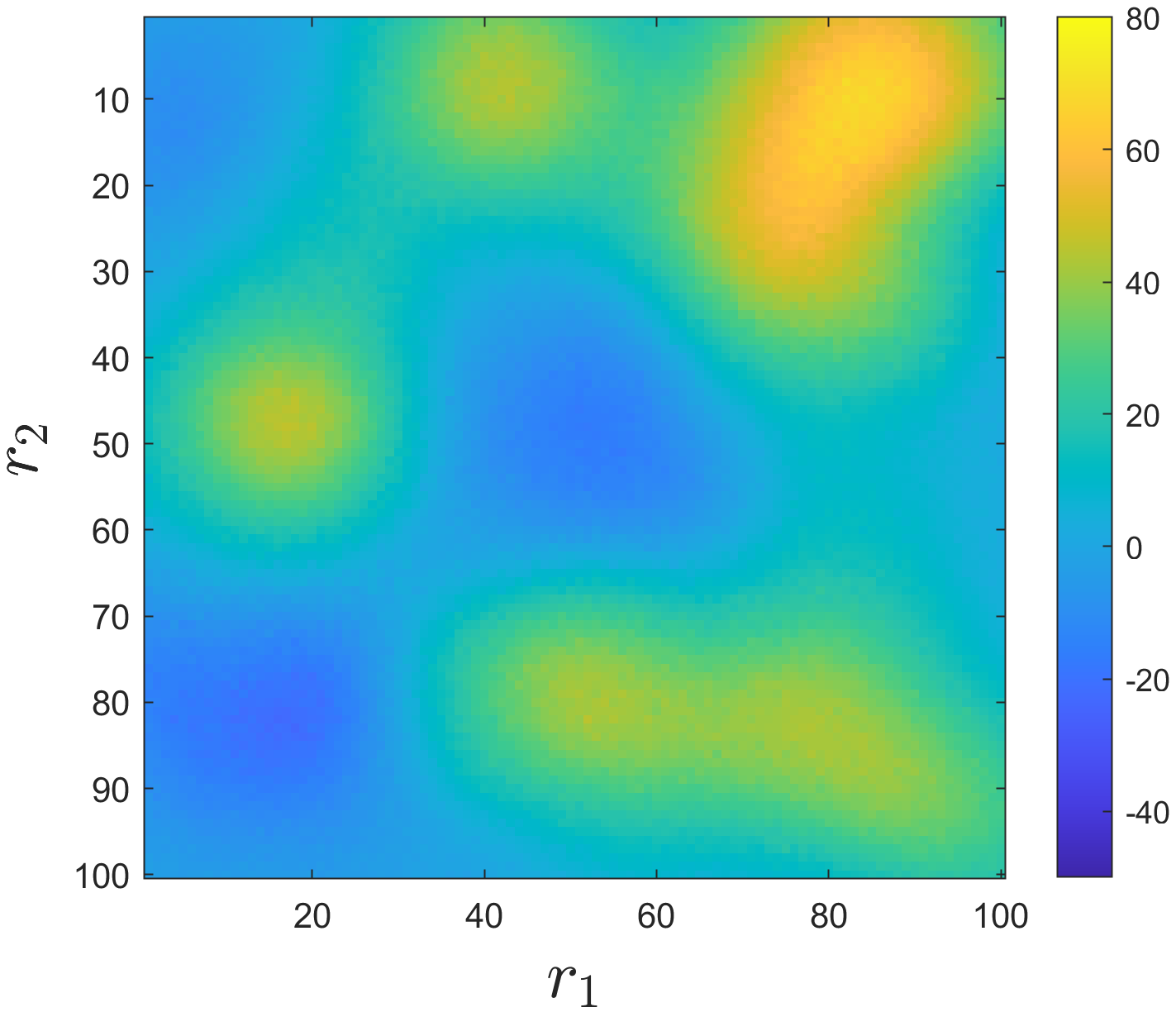}}
	\hspace{0.02\columnwidth}
	\subfigure[$t = 2$]{\includegraphics[width=0.46\columnwidth]{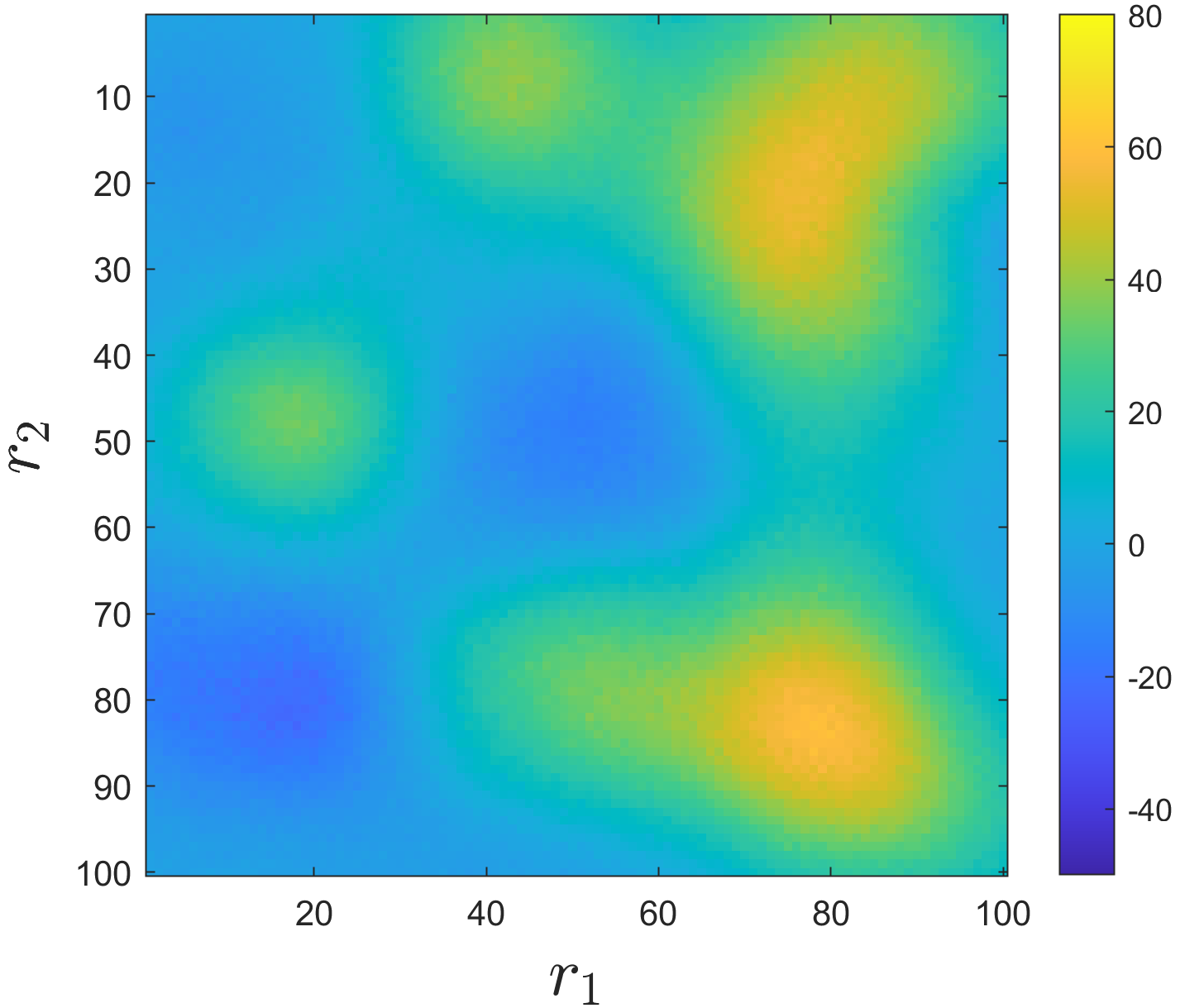}}
	\hspace{0.02\columnwidth}
	\subfigure[$t = 3$]{\includegraphics[width=0.46\columnwidth]{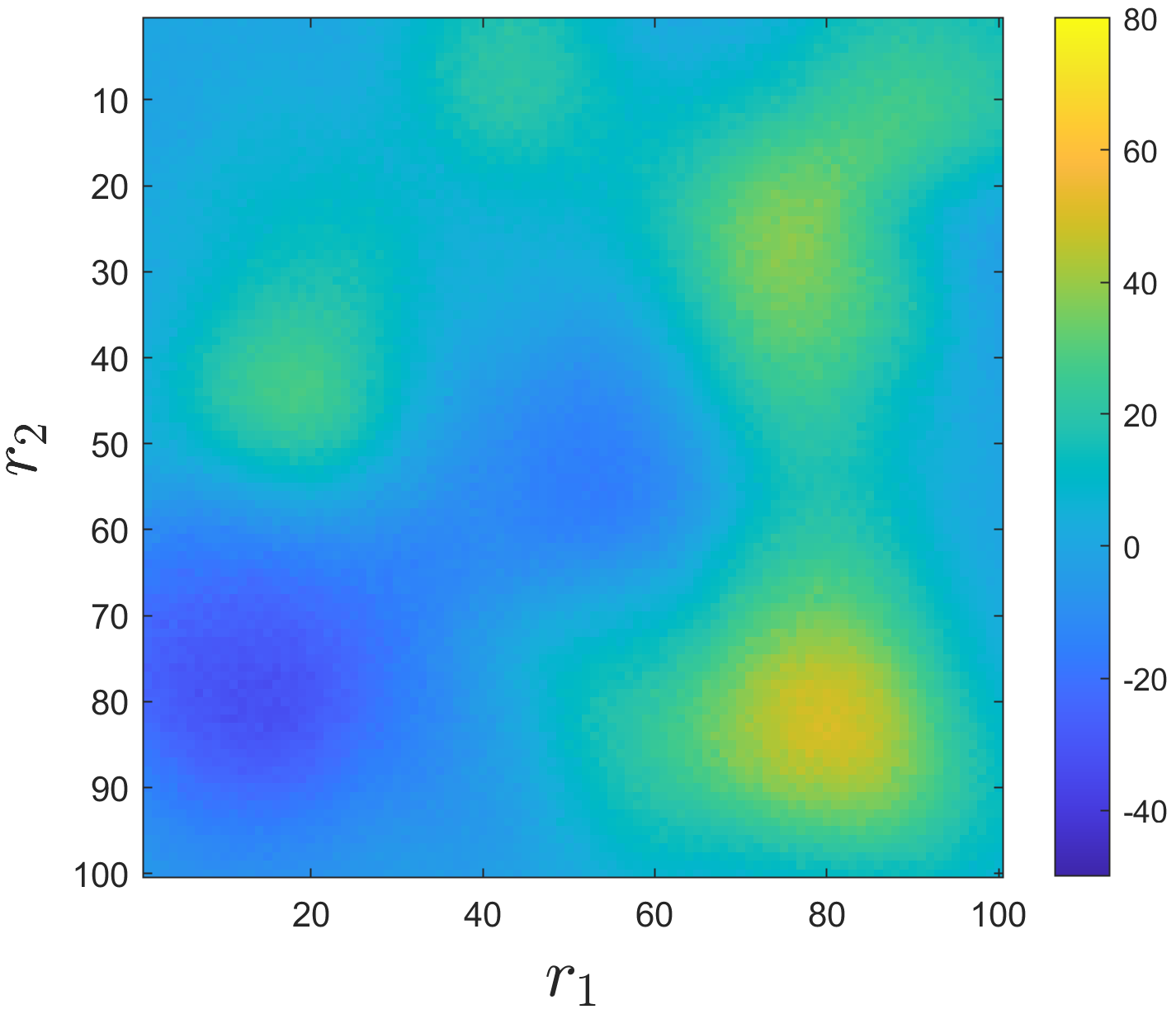}}
	\hspace{0.02\columnwidth}
	\subfigure[$t = 4$]{\includegraphics[width=0.46\columnwidth]{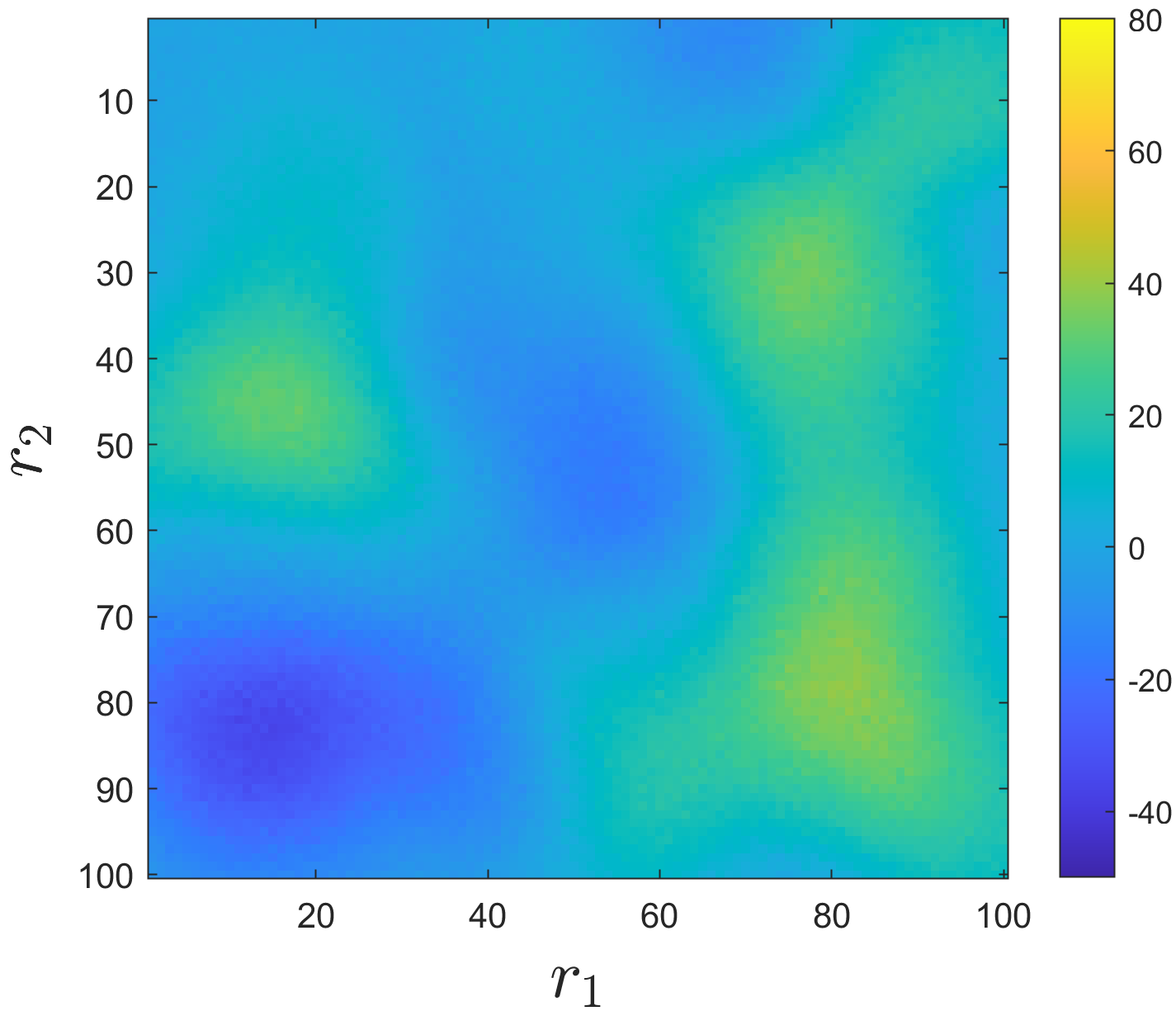}}
	\caption{S-VRNN generated samples for $\nData = 25.$}
	
	\label{fig-gen_threat_splitvrnn}
\end{figure}

\figfser{fig-gen_threat_svae}{fig-gen_threat_splitvrnn} 
show generated output samples for the three GAIMs with $\nData = 25$. 
The colorbar represents the intensity of the threat field. 
The samples generated by the S-VRNN adhere to the trend of
diminishing intensities of the threat field, as observed in 
\fig{fig-gen_threat_splitvrnn}. This is due to the $A$ matrix in 
\eqnnt{eq-dynamics} being Hurwitz, i.e., the coefficients $\threatState$
asymptotically settles to zero.
By contrast, the samples generated by the VRNN do not clearly follow this pattern. 
No discernible pattern can be seen in the S-VAE generated samples, which
are highly dissimilar from the training data.
The S-VAE samples exhibit a lack of diversity, as they appear clustered 
closely together in \figs{fig-metric_50}{fig-metric_25}, which is also
evident in \fig{fig-gen_threat_svae}.

We implemented the VRNN and S-VRNN for longer sequences of $T = 10$ time
steps and observed a similar pattern in the results. The S-VRNN
outputs remain statistically similar to the training data, whereas the VRNN 
outputs do not. For brevity, these results are not displayed here.


\begin{table*}
	\centering
	\caption{Statistical moments of the data samples for $\nData = 25.$}
	\label{tbl-moments-nd25}
	\resizebox{\textwidth}{!}{%
		\begin{tabular}{lrrr|rrr|rrr|rrr}
			\toprule
			\textbf{} & \multicolumn{3}{c|}{\textbf{Mean$(\mu)$}} & 
			\multicolumn{3}{c|}{\textbf{Variance$(\sigma^2)$}} & 
			\multicolumn{3}{c|}{\textbf{Skewness$(\gamma)$}} & 
			\multicolumn{3}{c}{\textbf{Kurtosis$(\kappa)$}} \\
			\midrule
			\textbf{} & $\mu_{\Sigma_{1}}$ & $\mu_{\Sigma_{2}}$ & $\mu_{\Sigma_{3}}$ & 
			$\sigma^2_{\Sigma_{1}}$ & $\sigma^2_{\Sigma_{2}}$ & $\sigma^2_{\Sigma_{3}}$ & 
			$\gamma_{\Sigma_{1}}$ & $\gamma_{\Sigma_{2}}$ & $\gamma_{\Sigma_{3}}$ & 
			$\kappa_{\Sigma_{1}}$ & $\kappa_{\Sigma_{2}}$ & $\kappa_{\Sigma_{3}}$ \\
			\midrule
			Training data pool & $-2.681\tento{3}$ & $9.65$ & $-11.81$ & $0.2381\tento{6}$ & $2.498\tento{6}$ 
			& 
			$2.153\tento{6}$ & 
			$0.17$ & $-0.07$ & $-0.39$ & $2.62$ & $2.61$ & $3.05$ \\
			S-VAE generated data & $-330.8$ & $22.13$ & $-9.67$ & $7491$ & $30.05\tento{3}$ & 
			$19.90\tento{3}$ & $-0.55$ & $-1.21$ & $-0.10$ & $1.99$ & $3.08$ & $2.46$ \\
			VRNN generated data & $3.168\tento{3}$ & $150.1$ & $-297.5$ & $0.9300\tento{6}$ & 
			$4.840\tento{6}$ & 
			$2.412\tento{6}$ & $0.10$ & $-0.81$ & $-0.24$ & $2.58$ & $2.50$ & $1.40$ \\
			S-VRNN generated data & $-2.935\tento{3}$ & $138.3$ & $-67.43$ & $0.8496\tento{6}$ & 
			$2.938\tento{6}$ & $2.375\tento{6}$ & $-0.30$ & $0.29$ & $0.39$ & $2.80$ & $2.03$ & $1.83$ \\
			\bottomrule
		\end{tabular}
	}
\end{table*}

\section{Conclusions}
\label{sec-conclusions}

In this paper, we developed a new deep learning model called the split 
variational recurrent neural network (S-VRNN) to address the problem of synthetic 
time-series data generation. The objective is to ensure similarity of the 
synthesized data to the training data. The S-VRNN is particularly suitable for
situations where the volume of training data is small, but some knowledge of 
the dynamics underlying the time-series data is available. 
The main innovation in this work is that 
we split the latent space of the S-VRNN into two subspaces.
The latent variables in one subspace are learned using the ``real-world'' data, 
whereas those in the other subspace are learned using the data as well as the
known underlying system dynamics. 
We conducted numerical experiments to compare the S-VRNN against two other
generative deep learning models: namely, a data-driven VRNN that learns from
data but does not consider the underlying dynamics, and a split variational
autoencoder that does consider the underlying dynamics but does not recognize
temporal dependencies. The results of our experiments showed that the VRNN and
S-VRNN both outperform the S-VAE in terms of statistical similarity of generated 
data to the trainig data. More importantly, we observed that the S-VRNN 
significantly outperforms the VRNN when the volume of training data is small.
This observation provides a promising basis for a longer-term investigation into
the S-VRNN for synthetic data generation in engineering applications where real-world
operational data are scarce.

\bibliographystyle{ieeetr}
\bibliography{References}

\end{document}